\pdfoutput=1 
\documentclass[sigconf, nonacm]{acmart}

\AtBeginDocument{%
  \providecommand\BibTeX{{
    \normalfont B\kern-0.5em{\scshape i\kern-0.25em b}\kern-0.8em\TeX}}}

\usepackage{subcaption}
\usepackage{amsmath,amsfonts}
\usepackage{graphicx}
\usepackage{textcomp}
\usepackage{xcolor}
\def\BibTeX{{\rm B\kern-.05em{\sc i\kern-.025em b}\kern-.08em
		T\kern-.1667em\lower.7ex\hbox{E}\kern-.125emX}}

\usepackage{microtype}
\usepackage{booktabs} 

\usepackage{tabularx}
\usepackage[utf8]{inputenc}
\usepackage[english]{babel}
\usepackage{array,multirow,graphicx}
\usepackage{float}

\usepackage[group-separator={,}]{siunitx}
\usepackage{etoolbox}
\usepackage{stackengine}

\usepackage{url}

\usepackage{algorithm}
\usepackage{balance}
\usepackage[noend]{algpseudocode}

\makeatletter
\algrenewcommand\ALG@beginalgorithmic{\small}
\makeatother

\usepackage{enumitem}
\setitemize{noitemsep,topsep=0pt,parsep=0pt,partopsep=0pt}

\DeclareMathOperator*{\argmax}{arg\,max}

\newtheorem{definition}{Definition}

\begin{document}

\title{Efficient Subspace Search in Data Streams}

\author{Edouard Fouché, Florian Kalinke, Klemens Böhm}

\affiliation{%
  \institution{Karlsruhe Institute of Technology (KIT)}
}
\email{{edouard.fouche, florian.kalinke, klemens.boehm}@kit.edu}
\orcid{0000-0003-0157-7648}


\begin{abstract}
In the real world, data streams are ubiquitous -- think of network traffic or sensor data. Mining patterns, e.g., outliers or clusters, from such data must take place in real time. This is challenging because~(1) streams often have high dimensionality, and (2) the data characteristics may change over time. Existing approaches tend to focus on only one aspect, either high dimensionality or the specifics of the streaming setting. For static data, a common approach to deal with high dimensionality -- known as subspace search -- extracts low-dimensional, `interesting' projections (subspaces), in which patterns are easier to find. 
In this paper, we address both Challenge (1) and (2) by generalising subspace search to data streams. Our approach, Streaming Greedy Maximum Random Deviation (SGMRD), monitors interesting subspaces in high-dimensional data streams. It leverages novel multivariate dependency estimators and monitoring techniques based on bandit theory. We show that the benefits of SGMRD are twofold: (i) It monitors subspaces efficiently, and (ii) this improves the results of downstream data mining tasks, such as outlier detection. Our experiments, performed against synthetic and real-world data, demonstrate that SGMRD outperforms its competitors by a large margin.
\end{abstract}

%
%

\settopmatter{printfolios=true} 
\maketitle
\thispagestyle{fancy}


\section{Introduction}

\subsection{Motivation}

Many sources generate streaming data: online advertising, transaction processing in banks, sensor networks, self-driving cars, twitter feeds, etc. Such streams have many dimensions, i.e., they are high-dimensional. 
Think of predictive maintenance. Here, data is seen as a stream of measurements from hundreds or thousands of sensors in a production plant. 
Extracting patterns in this setting is advantageous for many industrial applications and may lead to larger production volumes or reduce operational costs. 

A fundamental task of data analysis is to quantify the dependence between dimensions. 
This information helps understanding the data and often improves the result of subsequent tasks, e.g., outlier detection. With this in mind, researchers have proposed \textit{subspace search} methods, mainly for static data, to find interesting low-dimensional projections. 
Such projections tend to have much structure, i.e., high dependence among their dimensions. 
Subspace search is state-of-the-art to deal with data of high dimensionality and has numerous applications, including exploratory data mining \cite{DBLP:journals/sigkdd/AssentKMS07, DBLP:conf/ieeevast/TatuMFBSSK12}, outlier detection \cite{DBLP:conf/icde/ZhangGW08, DBLP:conf/icde/KellerMB12,  DBLP:journals/ijdsa/TrittenbachB19}, or clustering \cite{DBLP:conf/sigmod/ProcopiucJAM02, DBLP:conf/pkdd/KailingKKW03, DBLP:conf/icdm/BaumgartnerPKKK04, DBLP:conf/cikm/ParkL07, DBLP:conf/icdm/ZhangLW07}. 

One can see subspace search as an ensemble \textit{feature selection} method \cite{DBLP:journals/jmlr/GuyonE03}, as the goal is to find several projections (subspaces), not just one. 
The underlying assumption of subspace search is that patterns (e.g., outliers, clusters) may hide in various subspaces \cite{DBLP:books/sp/Aggarwal2013}, and that, when restricting the search to a single subspace, one may miss some patterns. 
In a nutshell, existing subspace search methods consist of two building blocks: 

(1) A \textit{quality measure} to quantify the `interestingness' of a subspace, i.e., the potential to reveal patterns. Intuitively, subspaces with `structure' are more likely to contain outliers or clusters \cite{DBLP:conf/icde/KellerMB12}. That measure often is a multivariate measure of dependence. 

(2) A \textit{search scheme} to explore the set of subspaces. Since this set grows exponentially with dimensionality, inspecting every subspace is not possible, and the search typically is a heuristic, i.e., a trade-off between completeness of the search and result quality.

These two items tend to be specific for a given data mining algorithm, e.g., a certain clustering method. 
The search then is helpful for this particular algorithm, but does not generalise beyond \cite{DBLP:conf/ieeevast/TatuMFBSSK12}. Next, existing methods for subspace search tend to assume static data. 
A straightforward generalisation to streams -- i.e., repeating the search periodically -- is computationally expensive and limited by the speed of new observations arriving. 

In this paper, we facilitate subspace search for streams. 
The core idea is to maintain a set of high-quality subspaces over time, by updating subspace-search results continuously. 
Existing approaches are much less efficient in practice, as we will show. 

\subsection{Challenges}
\label{challenges}

Searching for subspaces is difficult with static data already, because the number of subspaces increases exponentially with dimensionality. \cite{DBLP:books/sp/Aggarwal2013} compared the task of finding a pattern (e.g., an outlier) in high-dimensional spaces to that of searching for a needle in a haystack, while the haystack is one from an exponential number of haystacks. 
At the same time, \cite{DBLP:journals/ijdsa/TrittenbachB19} showed that to ensure high-quality results, the set of subspaces found must also be diverse, and that earlier methods yield subspaces with much redundancy.

The streaming setting comes as an additional but orthogonal challenge. Stream mining algorithms are complex and must satisfy several constraints \cite{doi:10.1198/1061860032544}, which we summarise as follows:

\textbf{C1: Efficiency.}  The algorithm must spend a short constant time and a constant amount of memory to process each record.

\textbf{C2: Single Scan.} The algorithm may perform at most one scan over the data -- no access to past observations. 

\textbf{C3: Adaptation.} Whenever the data distribution changes, the algorithm must adapt, e.g., by forgetting outdated information.

\textbf{C4: Anytime.} The algorithm results must be available at any point in time. The quality of those results must ideally be at least as high as the quality of the results from a static system. 

While a periodic recomputation of existing, static methods may cope with \textbf{C2} and \textbf{C3}, this is not efficient (\textbf{C1}). 
Additionally, results may not be available in an anytime fashion (\textbf{C4}). 

Considering these challenges, the analogy above becomes more complex: The haystacks (subspaces) of interest are not only hidden but also change over time, together with the location of the needle. 

\begin{figure}
	\centering
	\includegraphics[width=\linewidth]{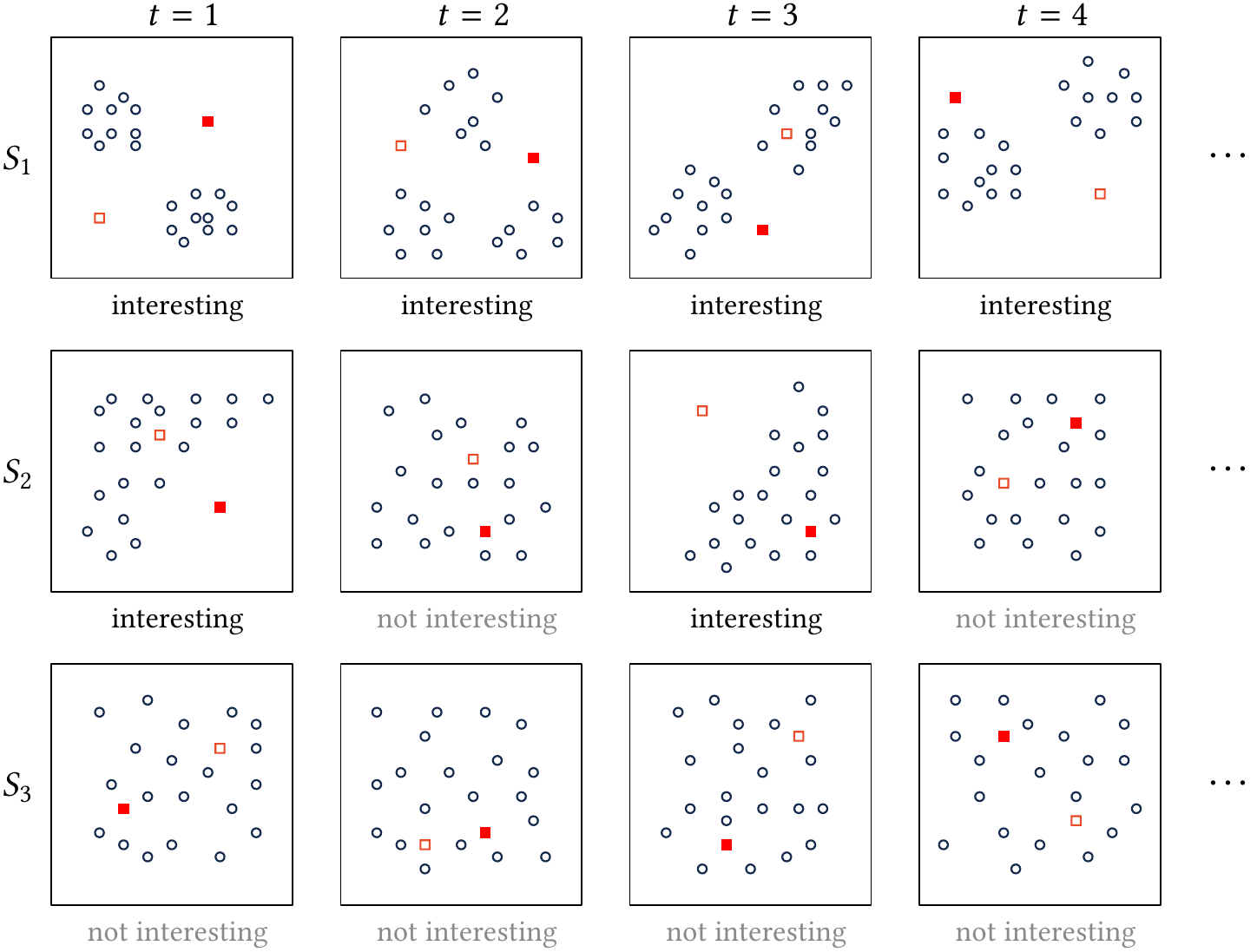}
	\caption{Subspace Search in Data Streams with outliers.} 
	\label{fig:outlier_example}
\end{figure}

Now, imagine that the needles of interest are the outliers in a data stream. In high-dimensional data streams, the challenge is to find the subspaces (`haystacks') in which outliers may be visible. We illustrate this idea in Figure \ref{fig:outlier_example}, with a fictitious data stream. Each column is a snapshot of the latest observations at time $t$ and each row represents a different 2-dimensional subspace $S_1, S_2, S_3$. The red squares are two outliers at each time step. As we can see, $S_1$ is interesting for $t \in \{1,2,3,4\}$, as its structure is such that it is prone to reveal outliers. In comparison, $S_2$ is only interesting at time $t = \{1,3\}$. In turn, $S_3$ does not help to reveal outliers, as the observations mostly are uniformly distributed in this subspace. The difficulty is that there is an exponential number of potentially interesting subspaces at any time. 
 
\subsection{Contributions}

\textbf{We facilitate subspace search in data streams}. After formulating the problem, we propose a new subspace search method, called Streaming Greedy Maximum Random Deviation (SGMRD). 

\textbf{We show that our method fulfils the above constraints}. Inspired by existing static methods \cite{DBLP:conf/icde/KellerMB12, DBLP:journals/ijdsa/TrittenbachB19}, SGMRD leverages a new multivariate dependency measure and Multi-Armed Bandit (MAB) algorithms to update the results of the search in data streams. 

\textbf{We perform extensive experiments}, based on an assortment of synthetic and real-world data. They show that (1) our approach leads to efficient monitoring of relevant subspaces, and (2) such monitoring also improves subsequent data mining tasks, such as outlier detection. We compare our approach to competitive baselines and state-of-the-art methods. 

\textbf{We release our source code, experiments and documentation on GitHub}\footnote{\url{https://github.com/edouardfouche/SGMRD}}, to help with the reproducibility of our study.

Outline: Section \ref{sec:related} covers related work, Section \ref{sec:notation} our notation. Section \ref{sec:problem} formulates the problem of subspace search in data streams. 
Section \ref{sec:subspace} presents our approach. 
Section \ref{sec:experiment} outlines the experimental setup. Section \ref{sec:results} presents our results. Section \ref{sec:conclusions} concludes. 

\section{Related Work}
\label{sec:related}

Many methods for subspace search exist, but almost all of them are either coupled to a specific data mining algorithm or are limited to the static setting. 
For example, various approaches for streams \cite{DBLP:conf/ideas/KontakiPM06, DBLP:conf/icdm/ZhangLW07, DBLP:conf/icde/ZhangGW08, DBLP:conf/icde/Aggarwal09a}
only tend to work with a given static algorithm. Other methods in turn \cite{DBLP:conf/icde/KellerMB12,  DBLP:conf/aaai/WangRNBMX17, 7806077, DBLP:journals/ijdsa/TrittenbachB19} decouple the search from the actual task, but none of them can handle streams. The existing work on subspace search mostly focuses on individual applications \cite{DBLP:journals/sigkdd/ParsonsHL04, DBLP:journals/tkdd/KriegelKZ09, DBLP:journals/sadm/ZimekSK12}, e.g., clustering or outlier detection, while `general-purpose' subspace search has received less attention. 

To our knowledge, there exist two proposals to extend subspace search to streams in a general way: HCP-StreamMiner \cite{vanea2012instant} and StreamHiCS \cite{becker2016concept}. 
But these approaches boil down to a periodic repetition of the procedure in \cite{DBLP:conf/icde/KellerMB12} on synopses of the stream. 
We will see that our method outperforms these approaches. 
Greedy Maximum Deviation (GMD) \cite{DBLP:journals/ijdsa/TrittenbachB19} is the approach most similar to ours. 
It uses a so-called contrast measure \cite{DBLP:conf/icde/KellerMB12} to quantify the interestingness of a given subspace and builds a set of subspaces via a greedy heuristic. 
However, GMD assumes static data.

Subspace search has been used in the past to improve the results of data mining tasks such as outlier detection \cite{DBLP:conf/icde/ZhangGW08, DBLP:conf/icde/KellerMB12,  DBLP:journals/ijdsa/TrittenbachB19}. The authors compare their results with full-space static outlier detectors. We perform an analogous evaluation in the streaming setting and compare our results against several baselines and state-of-the-art stream outlier detectors, such as xStream \cite{10.1145/3219819.3220107} and RS-Stream \cite{DBLP:journals/kais/SatheA18}. See \cite{DBLP:journals/tkde/GuptaGAH14} for a survey of outlier detection in streams.

\section{Notation}
\label{sec:notation}

A data stream is a set of dimensions $ {D}=\{s_1, \dots, s_{d}\}$ and an open list of observations $ {B} = (\vec{ {x}}_{1}, \vec{ {x}}_{2}, \dots)$, where $\vec{ {x}}_j$ with $j \in \mathbb{N}^+$ is a vector of  $d$ values, and we see a dimension $s_i = (x_1^i, x_2^i, \dots)$ with $i = \{1,\dots,d \}$ as an open list of numerical values.

Since the stream is virtually infinite, we use the sliding window model: At any time $ {t} \geq 1$, we only keep the $ {w}$ latest observations, $ {W_t} = \left(\vec{ {x}}_{ {t}- {w}+1}, \dots, \vec{ {x}}_{ {t}} \right)$. We call a subspace $ {S}$ a projection of the window $ {W_t}$ on $| {S}|$ dimensions, with ${S} \subseteq  {D}$ and $| {S}| \leq d$. $\mathcal{P}( {S})$ is the power set of ${S}$, i.e., the set of all dimension subsets. 
We assume, without loss of generality, that observations are equidistant in time. 

Window-based approaches are useful to overcome the constraints of streams, because they require a single scan of data ({\textbf{C2}}) and capture the most recent observations. Thus, algorithms based on such synopses can adapt ({\textbf{C3}}) \cite{DBLP:journals/pai/Gama12}. Note that one could easily adapt our method to accommodate other summarisation techniques, such as the landmark window or reservoir sampling \cite{DBLP:journals/pai/Gama12}.

\section{Problem Formulation}
\label{sec:problem}

\subsection{Dimension-based Subspace Search}

Subspace search in the static setting has already been formalised in the literature. The goal is to find a set of subspaces that fulfils a specific notion of optimality. Such subspaces must at the same time (1) be likely to reveal patterns (the `haystacks' from the analogy above) and (2) be diverse, i.e., have low redundancy with each other. 

To this end, the idea is to deem a set of subspaces optimal if adding or removing a subspace to/from this set makes the search results worse. 
To ensure diversity, the notion of optimality of each subspace must be tied to a specific dimension. 
This way, the resulting set may consist of the best subspaces w.r.t. each dimension, and each dimension is represented in this manner. This is the essence of what we call `dimension-based' search. 

\begin{figure}
	\centering
	\includegraphics[width=\linewidth]{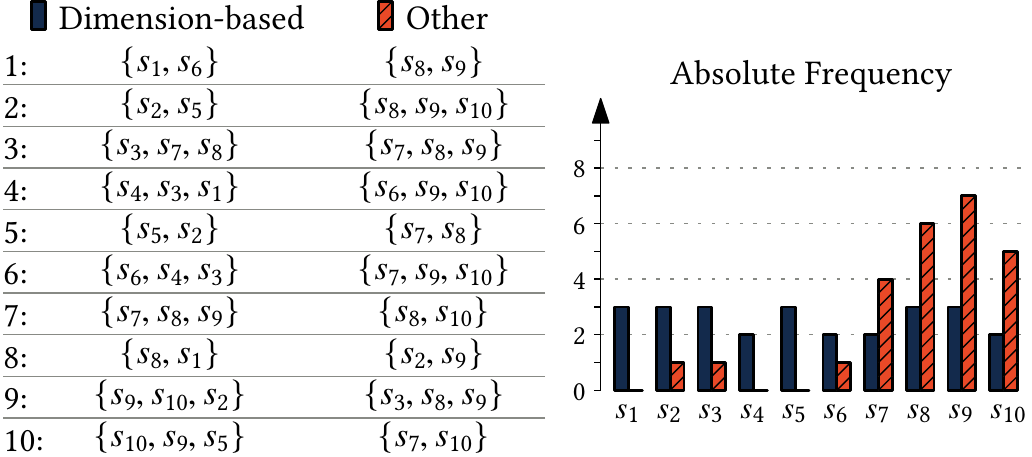}
	\caption{Dimension-based subspace search versus other methods on a toy data set ($d=10$).} 
	\label{fig:GMD_example}
\end{figure}

To illustrate this, we show in Figure \ref{fig:GMD_example} an exemplary result from a dimension-based search and from another scheme. Dimension-based results are more diverse compared to other results, which tend to over-represent some dimensions. Previous work \cite{DBLP:journals/ijdsa/TrittenbachB19} showed that the diversity from dimension-based approaches is key to improve the performance of subspace search algorithms. 

Formally, we define a so-called Dimension-Subspace Quality Function (D-SQF), to capture how much a dimension $s_i$ helps to reveal patterns in a subspace $S$: 
\begin{definition}[D-SQF]
	\label{def:sqf}
	For any $ {S} \in \mathcal{P}( {D})$ and any $ {s_i} \in  {D}$, a Dimension-Subspace Quality Function (D-SQF) is a function of type ${q}:   \mathcal{P}( {D}) \times  {D}  \mapsto [0,1]$ with $q(S, s_i) = 0, \forall  {s_i} \notin  {S}$.
\end{definition}
$q(S, s_i)=1$ means that $S$ has the maximum potential to reveal patterns w.r.t.\ $s_i$. Put differently, patterns may become more visible as one includes $s_i$ in $S$. This measure is asymmetric. For example, the quality of subspace $\{s_1, s_2\}$ does not need to be the same w.r.t.\ $s_1$ or $s_2$. In turn, $q(S, s_i)=0$ means that $S$ cannot reveal any patterns w.r.t.\ $s_i$. By definition, $q(S, s_i)=0$ if ${s_i}$ is not part of ${S}$ because the subspace cannot reveal any pattern in ${s_i}$. 

Recent studies \cite{DBLP:conf/aaai/WangRNBMX17, DBLP:journals/ijdsa/TrittenbachB19} instantiate such {D-SQF} as a measure of correlation, which one can estimate without any ex-post evaluation, i.e., it is not specific to any data mining algorithm. With this, subspace search remains independent of any downstream task. We can define a notion of subspace optimality: 
\begin{definition}[Optimal Subspace]
	A subspace $ {S} \in   \mathcal{P}( {D})$ is optimal w.r.t. $ {s_i} \in  {D}$ and a D-SQF $q$ if and only if 
	\begin{align*}
	{q}( {S},  {s_i}) \geq  {q}(S',  {s_i}) \quad \forall S' \in   \mathcal{P}( {D}).
	\end{align*}
\end{definition}
Then the optimal subspace set $\mathbb{S}^*$ is the set of optimal subspaces in $D$ w.r.t.\ each dimension $s_i \in D$:
\begin{definition}[Optimal Subspace Set]	\label{def:optimalset} A set $ \mathbb{S}^*$ is optimal w.r.t.\ $D$ and a D-SQF $ {q}$ if and only if
	\begin{align*}
	\forall s_i \in D,~ \exists S \in \mathbb{S}^* ~s.t.~ \forall S' \in \mathcal{P}(D),~ q(S,s_i) \geq q(S',s_i).
	\end{align*}
\end{definition}
Thus, the optimal set $\mathbb{S}^*$ contains one subspace $S$ for each dimension $ {s_i} \in {D}$, each one maximising the quality $ {q}$ w.r.t.\ $ {s_i}$. 
I.e., we can see $\mathbb{S}^*$ as a mapping of each dimension $ {s_i} \in {D}$ to an optimal subspace w.r.t.\ that dimension: 
\begin{align*}
\mathbb{S}^*:  {s_i} \in  {D} \mapsto  {S} \in   \mathcal{P}( {D}) \quad s.t. \quad \forall S' \in   \mathcal{P}( {D}) \quad  {q}({S}, s_i) \geq  {q}(S', s_i).
\end{align*}

Finding $\mathbb{S}^*$ is NP-hard \cite{DBLP:phd/dnb/Nguyen15d}. 
This is because one needs to assess the {D-SQF} of a set whose size grows exponentially with the number of dimensions. In fact, even finding a single optimal subspace is NP-hard. Thus, existing techniques do not guarantee the optimality of the results, but instead target at a good approximation of $\mathbb{S}^*$, while keeping the number of subspaces considered small. 

This idea is suitable in the static setting \cite{DBLP:conf/icde/KellerMB12, DBLP:conf/aaai/WangRNBMX17}. 
\cite{DBLP:journals/ijdsa/TrittenbachB19} showed that it leads to diverse sets of subspaces with better downstream mining results. 
Here, we propose a generalisation for streams. 

\subsection{Subspace Search in the Streaming Setting}
\label{sec:adaptation-stream}

\begin{figure}
	\centering
	\includegraphics[width=\linewidth]{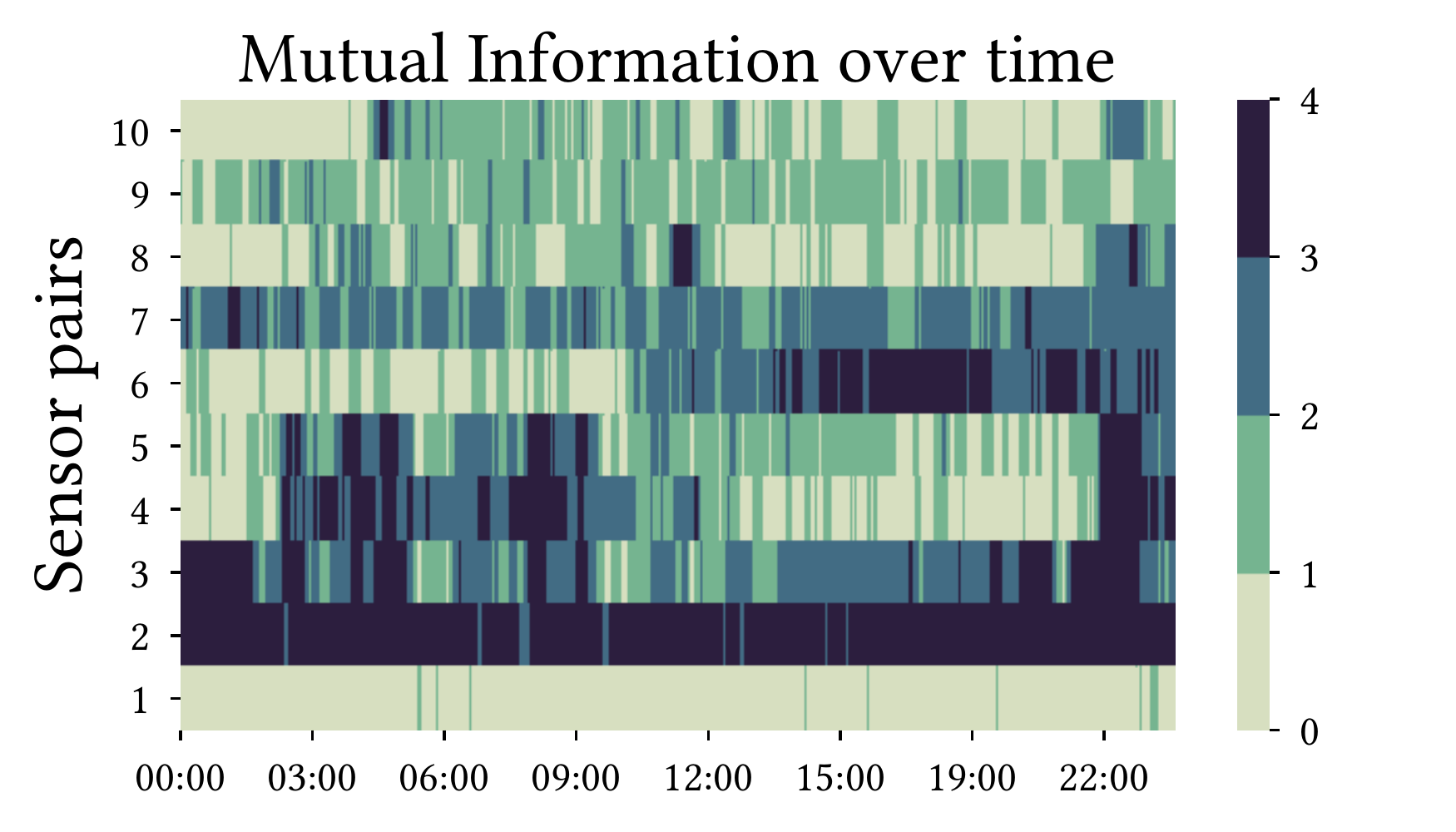}
	\caption{Evolution of Mutual Information between 10 sensor pairs in an experimental power plant for one day of data.} 
	\label{fig:MI_evolution}
\end{figure}

The quality of subspaces, estimated via statistical correlation measures, may change over time, manifesting a phenomenon known as `concept drift' \cite{DBLP:conf/ictai/BarddalGE15}. 
Thus, in streams, the quality function $q$ is time-dependent, and so we write  $\mathbb{S}_t^*$ and ${q_t}$. Then the problem becomes more complex --- observe the following example:

\begin{example}[Variation of Mutual Information]
	\label{example:mutual_information}
	We obtained measurement data from a power plant and computed the evolution  of correlation (estimated via Mutual Information) between a set of 10 sensor pairs for a single day. Figure \ref{fig:MI_evolution} graphs the results. The Mutual Information for pairs $1$ and $2$ remains stable for the whole duration, while it is more volatile for pairs $3$ to $6$. 
	The pairs $7$ to $10$ in turn show some change, but with less variance. 
\end{example}

As the example shows, some subspaces remain optimal for a longer period of time, while others frequently become sub-optimal. 
So the difficulty is that, even if one finds the set of subspaces $\mathbb{S}_t^*$, there is no guarantee that this set is optimal at time ${t}+1$. Next, it is impossible to even test the optimality of $\mathbb{S}_t^*$, because one would need to evaluate an exponential number of subspaces.

Let us assume that the cost of evaluating the quality of a subspace is constant across different subspaces and time. 
This is the case when considering existing correlation estimators. We define a function $\mathcal{E}_t: \mathcal{P}(D) \times D \mapsto \{0,1\}$ such that $\mathcal{E}_t(S, s_i) = 1$ if a given search scheme computes $q_t(S,s_i)$, else $0$. 
We can now formulate subspace search in data streams as a multi-objective optimisation problem at time ${t}$ with two conflicting objectives: 
\begin{enumerate}
	\item Find a set $\mathbb{S}_t$	which approximates $\mathbb{S}_t^*$ well. 
	I.e., minimise an objective $O_1 = \sum_{i=1}^{d} \left[q_t(\mathbb{S}_t^*({s_i}),s_i) - q_t(\mathbb{S}_t({s_i}),s_i)\right]$; the sum of the differences between the quality of the optimal set and the quality of the approximate set for each dimension. 
	\item Reduce the computation of the search, i.e., minimise an objective $O_2 = \sum_{{s_i}}^{{D}} \sum_{{S}}^{ \mathcal{P}({D})} \mathcal{E}_t(S, s_i)$; the number of subspaces for which one computes the quality at time $t$. 
\end{enumerate}
If $O_1 = 0$, then $\mathbb{S}_t \equiv \mathbb{S}_t^*$. Conversely, if $O_2 = 0$, then choosing $\mathbb{S}_t$ boils down to random guessing. Thus, $O_1$ and $O_2$ conflict. They capture the trade-off between the quality of the set of subspaces and the computation effort. 

Definition \ref{def:optimalset} implies that the search is independent for each dimension. Thus, to optimise $O_1$, we must find a dimension-based search algorithm, which for any dimension $s_i \in D$, returns a near-optimal subspace $S$ w.r.t.\ the dimension. 
More formally: 

\begin{definition}[Dimension-based Search Algorithm]
	A search algorithm, that is dimension-based, is a function $\mathit{Search}_t: D \mapsto \mathcal{P}(D)$, which for any dimension $s_i \in D$, returns a subspace $S$ minimising  $q_t(\mathbb{S}_t^*(s_i),s_i) - q_t(S,s_i)$ at time $t$. 
\end{definition}

Such an algorithm is associated with a cost, which depends on the number of subspaces evaluated. I.e., each run of $\mathit{Search}_t$ negatively impacts objective $O_2$. Thus, an additional challenge is to find an update policy $\pi: t \mapsto \mathcal{P}(D)$ to decide at any time for which dimension(s) one should repeat the search. We define it as follows: 

\begin{definition}[Update Policy]
	An update policy is a function \mbox{$\pi: t \mapsto \mathcal{P}(D)$.} 
\end{definition}
The policy returns for any time $t$ a set of dimensions $I_t \in \mathcal{P}(D)$, so that one repeats the $\mathit{Search}_t$ algorithm for $s_i \in I_t$.  

Overall, to achieve subspace search in data streams, we must come up with an adequate instantiation of the following elements: 

\begin{itemize}
	\item A D-SQF $q_t: \mathcal{P}( {D}) \times  {D}  \mapsto [0,1]$.
	\item A dimension-based search algorithm $\mathit{Search}_t: D \mapsto \mathcal{P}(D)$. 
	\item An update policy $\pi: t \mapsto \mathcal{P}(D)$.
\end{itemize}

Our approach, Streaming Greedy Maximum Random Deviation (SGMRD), addresses the challenges described previously by instantiating and combining each of these elements in a general framework. 

\section{Subspace Search in Data Streams}
\label{sec:subspace}

\begin{figure*} 
	\includegraphics[width=\linewidth]{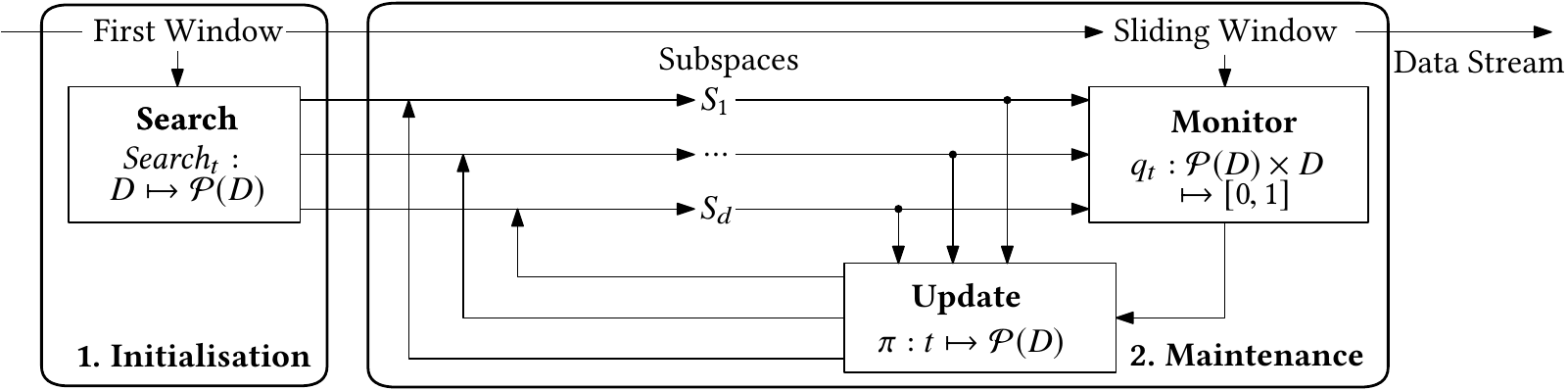}
	\caption{{SGMRD}: A High-Level Overview. At any time, SGMRD delivers a set of high-quality subspaces $S_1, \dots, S_d$.} 
	\label{fig:SGMRD_workflow}
\end{figure*} 

\subsection{Our Approach: {SGMRD}}

Figure~\ref{fig:SGMRD_workflow} shows an overview of our framework, in two steps: 
\begin{enumerate}[noitemsep]
	\item \textbf{Initialisation}. We find an initial set of subspaces $\mathbb{S}_0$, using the first ${w}$ observations in the stream. 
	To do so, we run the algorithm $\mathit{Search}_t$ for each dimension. 
	The outcome of this step is a set of subspaces $S_1, \dots, S_d$, one for each dimension. 
	\item \textbf{Maintenance}. For any new observation in the stream, we monitor the quality of subspaces $q_t(S, s_i)$ for $S \in \mathbb{S}_t$, and decide, by learning a policy $\pi$, for which dimension(s) we repeat the search, i.e., algorithm $\mathit{Search}_t$. 
\end{enumerate}

Our approach is general to some extent, as one could consider various instantiations of each building block (Search, Monitor and Update). With SGMRD, we instantiate the search function $\mathit{Search}_t$ as a greedy hill-climbing heuristic,  $q_t$ as an efficient dependency estimator and the policy $\pi$ as a Multi-Armed Bandit (MAB) algorithm with multiple plays. We describe the specifics of each block and explain our design decisions in the following sections. 

\subsubsection{Search}

\begin{figure}
	\centering
	\includegraphics[width=\linewidth]{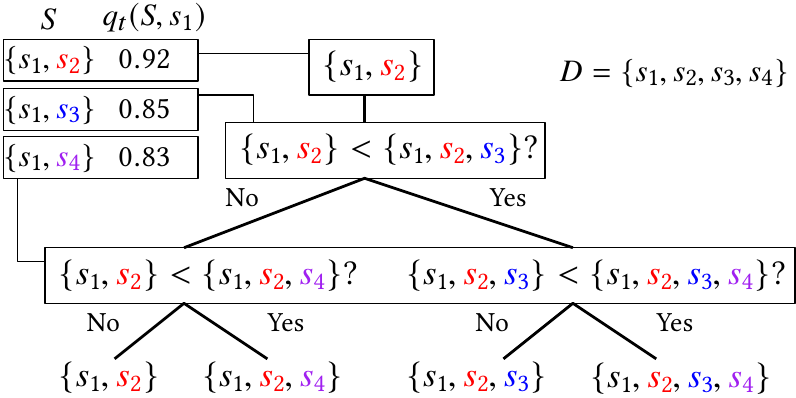}
	\caption{Example of search w.r.t. $s_1$ with four dimensions.} 
	\label{fig:SGMRD_gmd}
\end{figure}

{SGMRD}'s initialisation searches for an initial set of subspaces using the first observation window. Since finding the optimal set (cf.\ Definition~\ref{def:optimalset}) is not feasible, we instantiate $\mathit{Search}_t$ as a greedy hill-climbing heuristic. Our heuristic constructs subspaces in a bottom-up, greedy manner. 
Algorithm~\ref{alg:greedysearch} is our pseudo-code, and Figure \ref{fig:SGMRD_gmd} illustrates our idea with a toy example. 

In a first step (Line \ref{alg:greedysearch:line1}), we select the 2-dimensional subspace maximising the quality w.r.t.\ $s_i$. In our example, $s_i = s_1$, and the subspace with the highest quality is $\{s_1, s_2\}$. 
Then we iteratively test whether adding the dimension associated with the next best 2-dimensional subspace containing $s_i$ (Line \ref{alg:greedysearch:line4}) increases the quality of the current subspace (Line \ref{alg:greedysearch:line5}). If this is the case, we add it into the current subspace (Line \ref{alg:greedysearch:line6}), otherwise, we discard it. In our example, the heuristic first considers adding $s_3$, then $s_4$. 

The advantage of only considering 2-dimensional subspaces for selecting the next dimension is that it keeps the runtime of the search linear w.r.t.\ the number of dimensions. More precisely, we can see that the heuristic computes the quality of exactly $(d-1) + (d-2) = 2d -3$ subspaces. Thus, the runtime of Algorithm \ref{alg:greedysearch} is in $O(d)$. The search is independent for each dimension. At initialisation, we run it for each dimension, so the initialisation is in $O(d^2)$.

\begin{algorithm}
	\footnotesize
	\caption{$\mathit{Search}_t$(${s_i}$)}\label{alg:greedysearch} 
	\begin{algorithmic}[1]
		\Require A dimension ${s_i} \in {D}$ 
		
		\State $\mathcal{S}^{\mathit{max}} \gets {s_i} \cup \left (\argmax_{s_j \in {S}} q_t({s_i} \cup s_j, {s_i}) \right ) $ \label{alg:greedysearch:line1}
		\State ${S} \gets {S} \setminus \mathcal{S}^{\mathit{max}}$
		
		\While{ ${S}$ is not empty}
		\State $\mathcal{S}^{\mathit{cand}} \gets \argmax_{s_j \in {S}} q_t({s_i} \cup s_j, {s_i}) $ \label{alg:greedysearch:line4}
		\If{$q_t(\mathcal{S}^{\mathit{max}} \cup \mathcal{S}^{\mathit{cand}}, {s_i}) > q_t( \mathcal{S}^{\mathit{max}}, {s_i})$}  \label{alg:greedysearch:line5}
		\State $\mathcal{S}^{\mathit{max}} \gets \mathcal{S}^{\mathit{max}} \cup \mathcal{S}^{\mathit{cand}}$ \label{alg:greedysearch:line6}
		\EndIf
		\State ${S} \gets {S} \setminus \mathcal{S}^{\mathit{cand}}$
		\EndWhile
		
		\State \textbf{return} subspace $\mathcal{S}^{\mathit{max}} \subseteq {D}$
	\end{algorithmic}
\end{algorithm}

\subsubsection{Monitor}

So far, we did not discuss any concrete instantiation of the quality $q_t$. In practice, one has only a sample of observations, and thus, the quality can only be \textit{estimated} from a limited number of points. In what follows, we describe a new method to estimate the quality. Considering the constraints from the streaming setting (cf. Section \ref{challenges}) and the nature of the required quality function, our method must fulfil the following technical requirements: 

\textbf{Efficiency.} Since the search includes estimating the quality of numerous subspaces, the quality-estimation procedure must be efficient, to cope with the streaming constraints (\textbf{C1}).  

\textbf{Multivariate.} Since subspaces can have an arbitrary number of dimensions, the quality measure must be multivariate. Traditional dependency estimators in turn are bivariate  \cite{10.2307/2289859}.

\textbf{Asymmetric.} Since the quality values are specific for a given dimension, the measure is not symmetric. 

We define the quality as a measure of non-independence in subspace $S$ w.r.t.\ $s_i$. Observe the following definition of independence: 
\begin{definition}[Independence of Random Variables]
	A random vector $X=\{X_1, \dots, X_n\}$ is independent if and only if $p_X = \prod_{i=1}^{n} p_{X_i}$, where $p_X$ is the joint distribution and $p_{X_1}, \dots, p_{X_n}$ are the marginal distributions.
\end{definition}
This implies that the marginal distributions of each variable $X_i$ must be equal to their conditional distribution w.r.t.\ $X \setminus X_i$. By seeing each dimension as a random variable, we can estimate the quality as a degree of non-independence w.r.t.\ a dimension $X_i$, and quantify it as the discrepancy between the empirical marginal distribution $\hat{p}_{X_i}$ and the conditional distribution  $\hat{p}_{X_i|(X \setminus X_i)}$. 

For the ease of discussion, let us consider $q_t(\{s_1, s_2\}, s_1)$. Then,
\begin{align}
q_t(\{s_1, s_2\},s_1) \propto disc\left(\hat{p}_{s_1}, \hat{p}_{s_1|s_2}\right),
\end{align}
where $disc$ is the discrepancy between both distributions. 

To estimate this discrepancy, we propose a bootstrap method. 
Iteratively, we take a random condition w.r.t.\ the dimensions $S \setminus s_1$ (i.e., restricting the other dimensions to a random interval) and perform a statistical test between the sets of observations within and outside of this random condition w.r.t.\ $s_1$. 

\begin{figure}
	\hfill
	\begin{subfigure}{\linewidth}
		\begin{center}
			\includegraphics[width=\linewidth, trim=0 0.5cm 0 0]{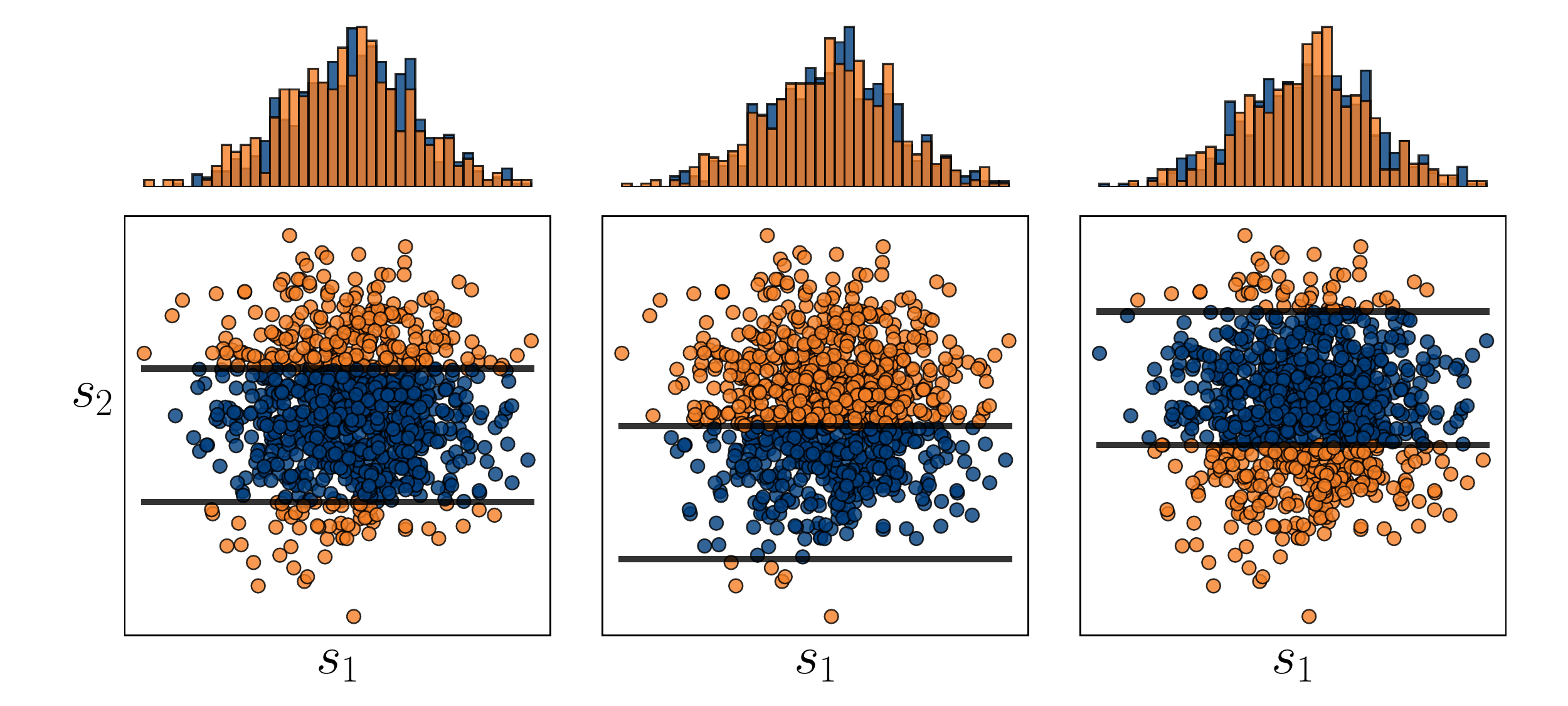}
		\end{center}
		\label{fig:slicing-gaussian}
	\end{subfigure}\par
	\begin{subfigure}{\linewidth}
		\begin{center}
			\includegraphics[width=\linewidth, trim=0 0.5cm 0 0]{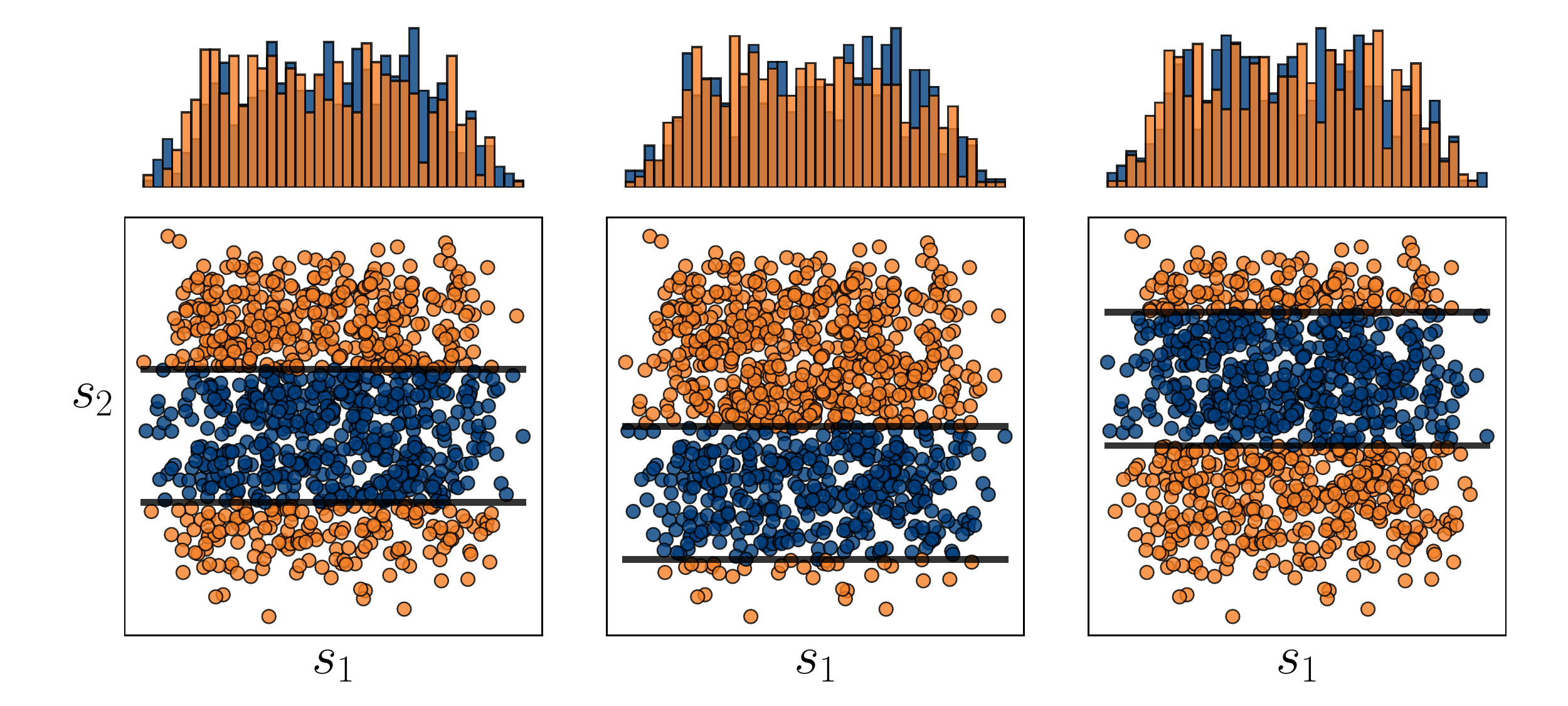}
		\end{center}
		\label{fig:slicing-independent}
	\end{subfigure}\par
	\begin{subfigure}{\linewidth}
		\begin{center}
			\includegraphics[width=\linewidth, trim=0 0.5cm 0 0]{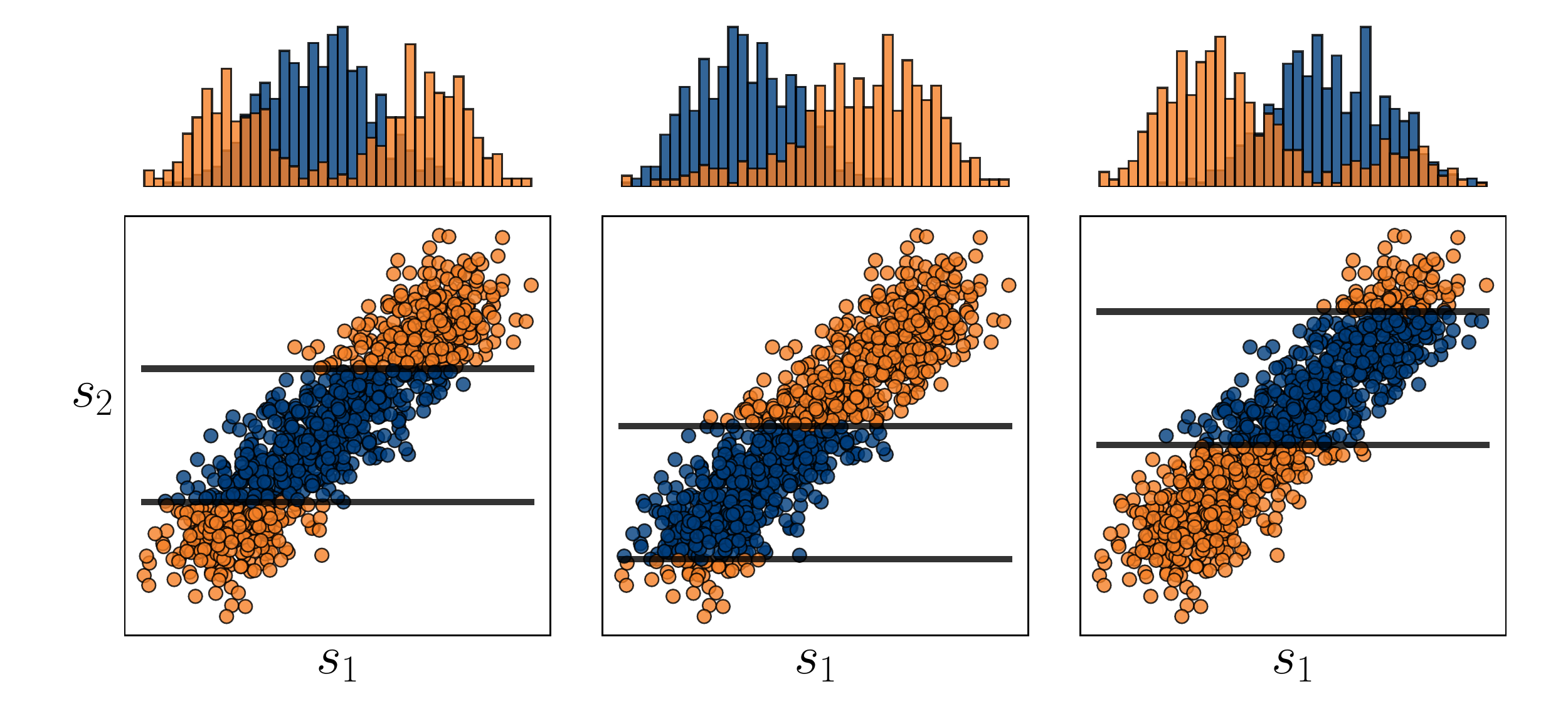}
		\end{center}
		\label{fig:slicing-linear}
	\end{subfigure}
	\begin{subfigure}{\linewidth}
		\begin{center}
			\includegraphics[width=\linewidth, trim=0 0.5cm 0 0]{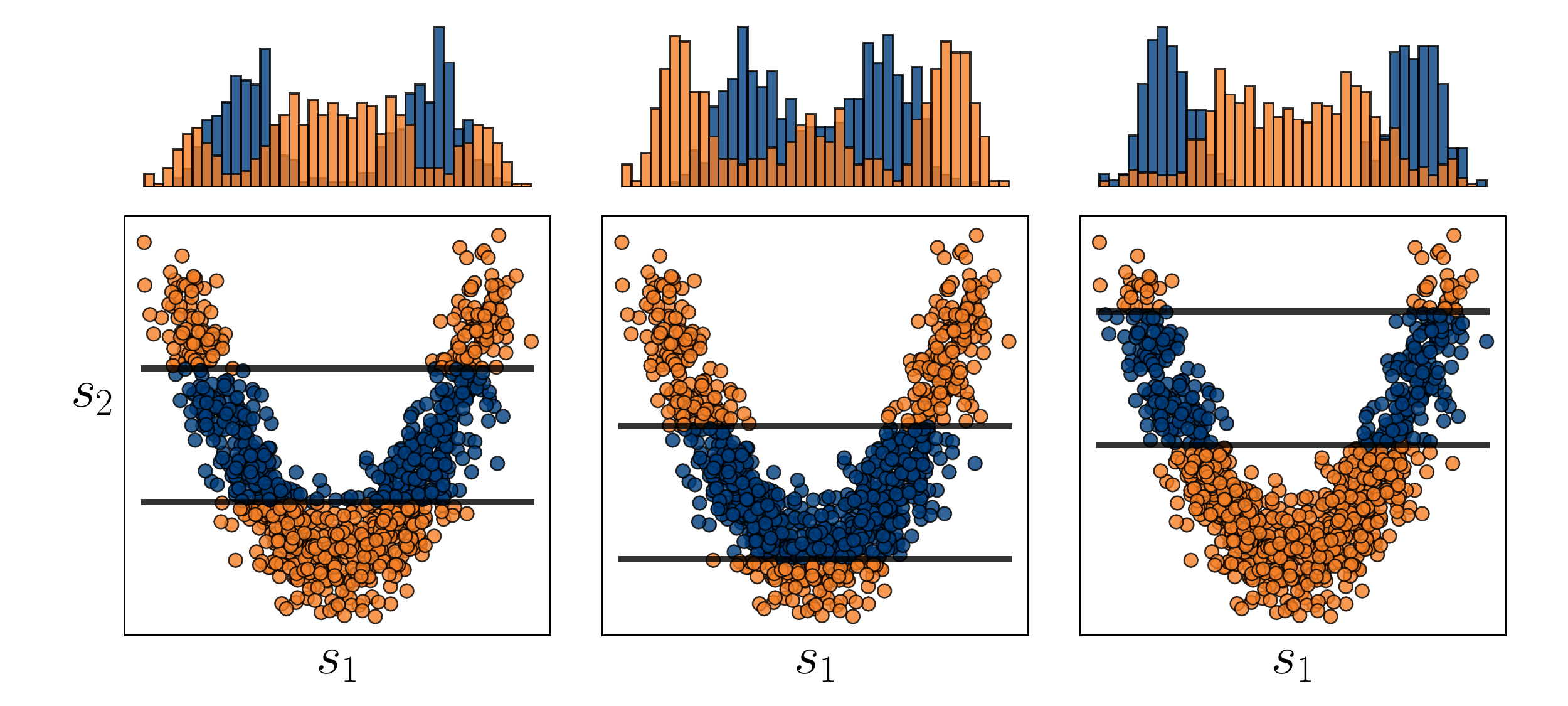}
		\end{center}
		\label{fig:slicing-circle}
	\end{subfigure}
	\caption{Our approach to estimate subspace quality.}
	\label{fig:slicing}
\end{figure}

Figure \ref{fig:slicing} illustrates this idea with two subspaces showing an independent 2-dimensional distribution and two subspaces with a linear and quadratic dependence. As we can see, over three randomly chosen conditions, the distributions of both sets of observations (see the histograms) are very similar for the first two subspaces, while they are markedly different for the other ones. Intuitively, subspaces with dependence are more likely to reveal patterns such as clusters or outliers, which are not visible in any other subspace. 

To estimate $q_t$, we average the discrepancies over  $M$ iterations: 
\begin{align}
q_t(S,s_i) = 1 - \frac{1}{M} \sum_{j=1}^{M} \mathcal{T}\left(\hat{p}_{S|c_j(s_i)}, \hat{p}_{S|\overline{c_j}(s_i)}\right),
\end{align}
where $c_j(s_i)$ is a random condition w.r.t.\ $s_i$, and the $\overline{c_j}(s_i)$ is its complementary condition in the current window of data. The resulting two sets of observations respectively are in dark blue and light orange in the figure. $\mathcal{T}$ yields the p-value of a two-sample Kolmogorov-Smirnov test. This test is adequate as it has high power and it is non-parametric. 
However, other tests are possible as well, see \cite{Siegel1956}. In the end, $q_t(S,s_i)$ converges to $1$ as the evidence against independence between both conditional distributions increases. It is efficient, because testing requires only linear time, while conditioning requires a one-time sorting of the elements in the window.

Our examples in Figure \ref{fig:slicing} are limited to two dimensions for the sake of illustration. However, one can easily extend this principle to multiple dimensions, as in \cite{DBLP:conf/icde/KellerMB12}. Based on Hoeffding's inequality \cite{doi:10.1080/01621459.1963.10500830}, it is easy to show that our measure
converges to its expected value as the number of iterations $M$ increases, formally:
\begin{align}
\Pr \left(|q_t(S,s_i) - \mathbb{E}(q_t(S,s_i))| \geq \varepsilon \right) \leq 2e^{-2M\varepsilon^2}.
\end{align}
See \cite{DBLP:conf/ssdbm/FoucheB19, DBLP:conf/ssdbm/FoucheBMKB20} for the corresponding proof. This brings several advantages: We can estimate the quality in quasi-linear time w.r.t.\ the number of points and reduce the number of iterations $M$ to adapt to the speed of streams. Thus, the approach is anytime (\textbf{C4}).

To monitor the quality estimates of each subspace in $\mathbb{S}_t$ over time, we smooth out the statistical fluctuations of the estimation process via exponential smoothing with parameter $\gamma$. Preliminary experiments showed that $M=100$ with $\gamma = 0.9$ leads to good estimation quality and performance w.r.t.\ downstream tasks. The outcome is a smoothed quality function $Q_t: {D}\mapsto [0,1]$:
\begin{align}
\label{eg:monitoring}
Q_{t+1}({s_i}) = \gamma \cdot q_t(\mathbb{S}_t(s_i), {s_i}) + (1- \gamma) \cdot {q}_{{t}+1}(\mathbb{S}_t(s_i), {s_i}).
\end{align}

\subsubsection{Update}
\label{sec:update}

The initialisation step is relatively expensive, because one needs to run the search (Algorithm \ref{alg:greedysearch}) for each dimension. 

While running the search for every dimension optimises the quality of the subspace set (Objective $O_1$), it curbs efficiency (Objective $O_2$). However, as in Example \ref{example:mutual_information}, the dependence between dimensions can unexpectedly change over time. Thus, without any assumption, it is not possible to exploit our knowledge from step $t-1$. We mitigate the computational cost over time by only repeating the search for a few dimensions at each time step.

The challenge is to find a policy $\pi: t \mapsto \mathcal{P}(D)$ to decide at any time step $t$ for which $L$ dimensions to repeat the search, where $L < d$ is a budget per time step. The budget $L$ can either be set by the user, or it is based on the computational time available between two subsequent observations. 

To solve this challenge, we cast the decisions of the policy $\pi$ as a MAB problem with multiple plays  \cite{DBLP:conf/icml/KomiyamaHN15}. Multi-Armed Bandit (MAB) models are useful tools to capture the trade-offs of sequential decision making problems. In what follows, we use the common notation from the bandit literature, as in \cite{DBLP:journals/ftml/BubeckC12}. We model each dimension $s_i \in D$ as an `arm'. In each round $t=\{1,\dots,T\}$, the policy $\pi$ selects $L < d$ arms $I_t \subset D$ and runs the search for each $s_i \in I_t$.  Then there are two possible outcomes for each $s_i$:  
\begin{enumerate}[noitemsep]
	\item \textbf{Success}: $\mathit{Search}_t(s_i)$ yields a better subspace than $\mathbb{S}_{t-1}(s_i)$, 
	so we update $\mathbb{S}_t({s_i})$. The policy receives a reward of $1$. 
	\item \textbf{Failure}: $\mathit{Search}_t(s_i)$ does not yield a better subspace, so we set $\mathbb{S}_t({s_i}) \gets \mathbb{S}_{{{t}}-1}({s_i})$. The reward is $0$. 
\end{enumerate}
Since the rewards are binary, we can associate each arm $s_i \in D$ with a Bernoulli distribution with unknown mean $\mu_i$. The goal of a multiple-play MAB algorithm is to find the top-$L$ arms with largest reward expectation $\mu_i$ from empirical observations with as few trials as possible \cite{DBLP:conf/icml/KomiyamaHN15}. In our case, this means finding the $L$ dimensions for which one must repeat the search more frequently. 

To find the best arms, we employ an algorithm known as Multiple-Play Thompson Sampling (MP-TS) \cite{DBLP:conf/icml/KomiyamaHN15}. In a nutshell, Thompson Sampling is a Bayesian inference heuristic which selects an arm based on posterior samples of the expectations of each arm. Since the Beta distribution is a conjugate prior for the Bernoulli distribution, $\mathit{Beta}(\alpha_i, \beta_i)$ is the prior belief for arm $i$, and for each arm $i$, we maintain a pair of parameters $\alpha_i, \beta_i$. At each step, we sample from each $\mathit{Beta}(\alpha_i, \beta_i)$ distribution and play the arms with highest value. Whenever we observe a success after playing arm $i$, we increment $\alpha_i$, otherwise we increment $\beta_i$. Assuming a uniform initial prior, we initialise $\alpha_i, \beta_i = 1$. The idea of Thompson Sampling traces back to 1933 \cite{Thompson1933}, but recent studies demonstrate its superiority over other bandits in theory and practice \cite{DBLP:conf/alt/KaufmannKM12, DBLP:conf/nips/ChapelleL11}. 

Intuitively, the behaviour of the policy $\pi$ is to explore in the beginning, by playing various arms, and then to exploit the high-reward arms, i.e., repeating the search for dimensions whose optimal subspace changes more frequently. Algorithm~\ref{SGMRD-Update} is the pseudo-code for our update step. 

\begin{algorithm}[ht]
	\footnotesize
	\caption{$\mathit{Update}_t$($\mathbb{S}_t$, $L$)}\label{SGMRD-Update} 
	\begin{algorithmic}[1]
		\Require A set of subspaces $\mathbb{S}_t$, the number of plays per round ${L}$.
		\For{$i = 1,\dots, {K}$}
		\State $\theta_i(t) \sim \mathit{Beta}(\alpha_i(t), \beta_i(t))$
		\EndFor 
		\State $ {I_t} = \argmax_{K' \subset [ {K}], |K'| = {L}} \sum_{i}^{K'}\theta_i(t)$ 
		\For{$i \in  {I_t}$}
		\State $S \gets \mathit{Search}_t({s_i})$  \Comment{Searching for a new subspace w.r.t. $s_i$}
		\If{${q}_{{t}+1}({S}, {s_i}) > Q_{t+1}(\mathbb{S}_t({s_i}), {s_i})$ and ${S} \neq \mathbb{S}_t({s_i})$}  
		\State $\mathbb{S}_{{{t}}+1}({s_i}) \gets {S}$ ;  $\alpha_i(t+1) = \alpha_i(t) + 1$ 
		\Else 
		\State $\mathbb{S}_{{{t}}+1}({s_i}) \gets \mathbb{S}_t({s_i})$ ; $\beta_i(t+1) = \beta_i(t) + 1$ 
		\EndIf
		\EndFor
		\State \textbf{return} $\mathbb{S}_{{{t}}+1}$
	\end{algorithmic}
	
\end{algorithm}

\begin{algorithm}
	\footnotesize
	\caption{\textsc{{SGMRD}}((${D},{B}$), ${w}$, $L$, $\gamma$)}\label{alg:sgmrd} 
	\begin{algorithmic}[1]
		\Require Data stream (${D},{B}$), window size $\mathit{{w}} > 0$, $L>0$, $\gamma \in (0,1)$ 
		\State ${t} \gets {w}$ \Comment{\textbf{1. Initialisation}}
		\For{$s_i \in D$}
		\State $\mathbb{S}_t(s_i) \gets \mathit{Search}_t({s_i})$ \label{line:initialsearch} \Comment{Search}
		\EndFor
		
		\While{${B}$ has a new observation $\vec{{x}}_{{{t}}+1}$}   \Comment{\textbf{2. Maintenance}}
		\State ${t} \gets {t} + 1$
		
		\For{$s_i \in D$} \label{line:monitor} \Comment{Monitor}
		\State $Q_{t}({s_i}) = \gamma \cdot q_{t-1}(\mathbb{S}_{t-1}(s_i), {s_i}) + (1- \gamma) \cdot {q}_{{t}}(\mathbb{S}_t(s_i), {s_i})$ 
		\EndFor
		
		\State $\mathbb{S}_t \gets \mathit{Update}_t(\mathbb{S}_{{{t}}-1}, L)$ \label{line:update} \Comment{Update} 
		
		\EndWhile	
		\State \textbf{return at anytime} the set of subspaces $\mathbb{S}_t$ 
	\end{algorithmic}
\end{algorithm}

\subsubsection{{SGMRD}}

Algorithm \ref{alg:sgmrd} summarises our approach as pseudo-code. {SGMRD} finds an initial set of subspaces using the first window (Line \ref{line:initialsearch}). 
Then SGMRD monitors and updates the set of subspaces (Line \ref{line:monitor} to \ref{line:update}) for each new observation $\vec{x}_{t+1}$.

Overall, our method, SGMRD, is efficient ({\textbf{C1}}), as our quality estimates can be computed in linear time. It also requires a single scan of the data ({\textbf{C2}}), as we monitor subspaces over a sliding window. 
By design, SGMRD adapts ({\textbf{C3}}) to the environment by updating the subspace search results with parsimonious resource consumption, and our experiments will confirm this. Finally, results are available at any point in time ({\textbf{C4}}). 

In addition to reducing the number of plays ${L}$, one can further bring down the computational requirements of SGMRD by performing the update step (Line \ref{line:update}) only once every $v$ new observations. For simplicity, we have described Algorithm \ref{alg:sgmrd} with $v=1$.  

In our experiments, we will study the trade-off between the quality of the results and the cost associated with our method. 

\subsection{Downstream Data Mining}
\label{sec:downstream}

High-quality subspaces can be useful for virtually any downstream data mining task. For example, previous work \cite{DBLP:conf/icde/KellerMB12,  DBLP:conf/sdm/NguyenMV16, DBLP:conf/aaai/WangRNBMX17, DBLP:journals/ijdsa/TrittenbachB19} leverages subspaces to build ensemble-like outlier detectors. Other data mining tasks are possible as well, such as clustering \cite{ DBLP:conf/cikm/ParkL07, DBLP:conf/icdm/ZhangLW07, DBLP:journals/datamine/AgrawalGGR05, DBLP:conf/icde/Aggarwal09a, DBLP:conf/ideas/KontakiPM06}. Our approach, {SGMRD}, yields a set of subspaces $\mathbb{S}_t$ at any time. While our approach is not tied to any specific data mining task, we use outlier detection as an exemplary task in our evaluation. 

The articles just cited apply an outlier detector to each subspace in $\mathbb{S}_t$, and the final score is a combination of the individual scores. 
The outcome is a ranking of objects by decreasing `outlierness'. 
In the experiments that follow, we use the {LOF} detector
because it is a common baseline in the outlier detection literature \cite{DBLP:conf/icde/KellerMB12}. 

The best combination of the individual scores depends on the concrete application.
Literature has discussed this extensively  \cite{DBLP:journals/sigkdd/AggarwalS15, DBLP:journals/datamine/VinhCRBLRP16}. 
Several studies \cite{lazarevic2005feature, DBLP:conf/cidm/PokrajacLL07} argue that the average of the scores with {LOF} yields the best results overall in the static case, so we stick to this choice. The final outlier score of $\vec{{x}}_{{t}}$ is the average of the scores from each subspace across every window containing $\vec{{x}}_{{t}}$: 
\begin{align}
\textit{score}^{~}(\vec{{x}}_{{t}}) = \frac{1}{{w} \cdot d} \sum_{i=0}^{{w}} \sum_{{s_i}}^{\mathbb{S}_{{{t}}-i}} score_{W_{{{t}}-i}}^{\mathbb{S}_{{{t}}-i}({s_i})}(\vec{{x}}_{{t}}).
\end{align}
Again, one may also reduce the computation effort of outlier detection by evaluating the scores only once every $v$ time steps. 

\section{Experiment Setup}
\label{sec:experiment}

We evaluate the performance of our approach w.r.t.\ two aspects: (1)~the quality of subspace monitoring, i.e., how efficiently and effectively can {SGMRD} maintain a set of high-quality subspaces over time, and (2)~the benefits w.r.t.\ outlier detection, as an exemplary downstream data mining task.

\subsection{Evaluation Measures}

\subsubsection{Subspace Monitoring}

Evaluating the quality of subspace monitoring is difficult because finding $\mathbb{S}_t^*$ is computationally infeasible for non-trivial data. Thus, based on the definition of objective $O_1$ (cf. Section \ref{sec:adaptation-stream}), we measure the regret and the average quality: 
\begin{align}
R_{{T}} = \frac{1}{d} \sum_{{t}=0}^{{{T}}} \sum_{{i}=1}^{d} \left[{Q}_{{t}}^*(s_i) - Q_{{t}}(s_i) \right],~ \overline{Q}_{{t}} = \frac{1}{d}  \sum_{{i}=1}^{d} Q_{{t}}(s_i),
\end{align}
where $Q_{{t}}(s_i)$ is the quality w.r.t.\ $s_i$ as defined in Equation~\ref{eg:monitoring}, and ${Q}_{{t}}^*(s_i)$ is the quality obtained when one always updates the corresponding subspace, i.e., it the same as repeating the initialisation. $R_{{T}}$ is the regret, defined as the sum of the differences between ${Q}_{{t}}^*(s_i)$ and $Q_{{t}}(s_i)$ up to time $T$. $\overline{Q}_{{t}}$ is the average quality at time $t$, and $\overline{Q}_{{T}}$ is the average quality up to time $T$. 

To characterise the behaviour of our update strategy, we also look at the relative update frequency $F_{{T}}({s_i})$ for each dimension ${s_i} \in {D}$ and the rate of successful updates $U_{{T}}$, as in Section \ref{sec:update}, i.e.:
\begin{align}
F_{{T}}({s_i}) = \sum_{{t}=0}^{{{T}}} \frac{\mathbf{1}\left[{s_i} \in {I_t}\right]}{T},~ U_{{T}} = \sum_{{t}=0}^{{{T}}} \sum_{{i=1}}^{{d}} \frac{ \mathbf{1}\left[A_t(s_i) \wedge B_t(s_i)\right]}{d\cdot {{T}}}, \label{eq:U_T}
\end{align}
where ${s_i} \in {I_t}$ means that {SGMRD} has selected dimension ${s_i}$ at time ${t}$, and $A_t(s_i)$ and $B_t(s_i)$ are the conditions capturing whether the search was successful (i.e., the new subspace is different and of higher quality than the previous one): 
\begin{align}
A_t(s_i) &= \mathbb{S}_{{{t}}-1}({s_i}) \neq \mathbb{S}_t({s_i}), \\
B_t(s_i) &= Q_t(\mathbb{S}_{{{t}}-1}({s_i}), {s_i}) < q_t(\mathbb{S}_t({s_i}), {s_i}).
\end{align}

\subsubsection{Outlier Detection}

By definition, outliers are rare, so the detection of outliers is an imbalanced classification problem. We report the area under the ROC curve ({AUC}) and the Average Precision (AP), which are popular measures for evaluating outlier detection algorithms. 
Outlier detectors typically ranks the observation by decreasing `outlierness', measured as a score. In most applications, end users only check the top X\% items. 
So we report the Recall (R) and Precision (P) within the top X\% instances, with $\text{X} \in \{1,2,5\}$. 

\subsection{Data Sets}

We use an assortment of data sets for our evaluation. 
One is a data set from a real-world use case, corresponding to measurements in a pyrolysis plant. We also include four real-world data sets with outlier ground truth: \textsc{KDDCup99}, \textsc{Activity}, \textsc{Backblaze}, \textsc{Credit}. While the former two have frequently been used in the literature, the latter two are our own addition to this benchmark. 
To cope with the lack of publicly available data sets for outlier detection in the streaming setting, we also generate three synthetic benchmark data sets: \textsc{Synth10}, \textsc{Synth20} and \textsc{Synth50}. 
We describe each data set in detail hereafter; Table \ref{tab:databenchmark} summarises their main characteristics. 

\begin{table}[ht]
	\caption{Characteristics of the Benchmark Data Sets.}
	\label{tab:databenchmark}
	\addtolength{\tabcolsep}{-2pt}
	\centering
	\renewcommand{\arraystretch}{0.80}
	\begin{tabularx}{0.8\columnwidth}{l|ccc}
		Benchmark        & \# Instances & \# Dimensions & \% Outliers \\ \toprule
		\textsc{Pyro}  & \num{10000}      & 100       & NA                      \\ 
		\textsc{KDDCup99}  & \num{25000}      & 38       & 7.12                    \\ 
		\textsc{Activity}  & \num{22253}      & 51       & 10                     \\
		\textsc{Backblaze}  & \num{12600}      & 44       & 1                     \\
		\textsc{Credit}  & \num{284807}      & 29       & 0.17                     \\ \midrule
		\textsc{Synth10}  & \num{10000}      & 10       & 0.86                 \\ 
		\textsc{Synth20} & \num{10000}      & 20       & 0.88               \\ 
		\textsc{Synth50}  & \num{10000}      & 50       & 0.81                 \\ \bottomrule
	\end{tabularx}
\end{table}

\subsubsection{Real-World Data Sets}

\begin{itemize}[noitemsep]
	\item \textsc{Pyro:} This data set contains \num{10000} measurements (one per second) from a selection of 100 sensors, such as temperature or pressure, in various components of a pyrolysis plant. 
	We use this data set to evaluate how well our method can search for subspaces in data streams. 
	However, there is no ground truth, so we cannot use this data set for our downstream data mining application. 
	\item \textsc{Activity:} This data set, initially proposed in \cite{DBLP:conf/iswc/ReissS12}, describes different subjects performing various activities (e.g., walking, running), monitored via body-mounted sensors. Analogously to \cite{DBLP:journals/kais/SatheA18}, we took the \textit{walking data} of a single subject and replaced 10\% of the data with \textit{nordic walking} data, which we marked as outliers. 
	The rest of the elements are inliers. We obtained the original data set from \cite{Dua:2019}. 
	\item \textsc{KDDCup99:} This data set was part of the KDD Cup Challenge 1999. It is a network intrusion data set. Analogously to \cite{DBLP:journals/kais/SatheA18}, we excluded DDoS (Denial-of-Service) attacks and marked all other attacks as outliers. 
	We take a contiguous subset of \num{25000} data points. 
	We obtained this data set from \cite{Dua:2019}.
	\item \textsc{Backblaze:} Similarly as in \cite{DBLP:conf/spects/CPVCL17}, we obtained hard drive failure stats data from Backblaze\footnote{\url{https://www.backblaze.com/b2/hard-drive-test-data.html}}, a computer backup and cloud storage service. We prepare a benchmark with the data from Q1 2020. We select the hard drive model ST12000NM0007, because of its relatively high failure rate, and downsample the instances from the normal class such that the outliers represent $1\%$ of the total observations.  
	\item \textsc{Credit:} This data set was released via the `credit card fraud detection' challenge\footnote{\url{https://www.kaggle.com/mlg-ulb/creditcardfraud}} \cite{DBLP:journals/tnn/PozzoloBCAB18}. It is highly imbalanced, with only $0.17\%$ outliers. Also, it has comparably more instances. 
\end{itemize}

\subsubsection{Synthetic Benchmark Generation}

In addition, we create three data sets simulating \textit{concept drift} \cite{DBLP:conf/ictai/BarddalGE15} via random variations of the data distribution over time. We generate $n+1$ distributions $\Gamma_0, \Gamma_1, \dots, \Gamma_{n}$, and sample from each distribution $\Gamma_i$ a number $e$ of observations, while letting the distribution $\Gamma_i$ gradually drift towards the distribution $\Gamma_{i+1}$ as we sample from it. 

\begin{figure}
	\centering
	\includegraphics[width=\linewidth]{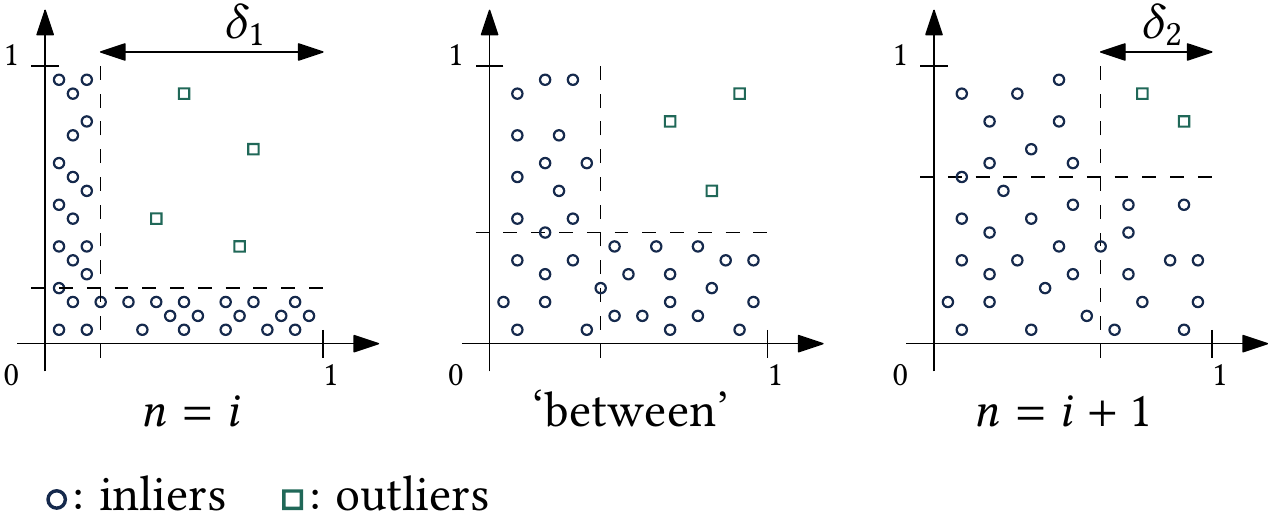}
	\caption{The synthetic benchmark generation process.}
	\label{fig:SGMRD_generation}
\end{figure}

We initialise $\Gamma_0 \equiv \mathcal{U}[0,1]$ over the full space $D$. Then we select a set of distinct subspaces (i.e., the subspaces do not have any dimension in common) from $ \mathcal{P}({D})$ for each other distribution, so that $50\%$ of the subspaces change from one distribution to the next one. 
For each subspace and, with a small probability $p \in (0,1)$, we sample the next point from $\mathcal{U}[\delta, 1]$ for each dimension with $\delta \in (0,1)$ chosen randomly. We call this point an `outlier'. With probability $1-p$, we sample the next point uniformly from the rest of the unit hypercube -- this is an `inlier'. Figure \ref{fig:SGMRD_generation} illustrates this principle with two dimensions. For the dimensions not part of any subspace, every observations are i.i.d.\ in $\mathcal{U}[0,1]$.

Outliers placed this way are said to be `non-trivial' \cite{DBLP:conf/icde/KellerMB12}, since they do not appear in any other subspace --- they are `hidden' in the data. Our goal is to evaluate to what extent different approaches can detect such outliers. 

We generate three benchmark data sets with $n=10$ distributions and $e=1000$ observations. We set $p < 0.01$, since outliers are rare by definition. \textsc{Synth10}, \textsc{Synth20}, \textsc{Synth50} have $10$, $20$ and $50$ dimensions respectively, with subspaces up to $5$ dimensions. 

Note that we release the code for our data generator and our real-world benchmark data sets via our GitHub repository as well. 

\subsection{Baselines and Competitors}

\subsubsection{Subspace Monitoring} 

Our goal is to assess how effectively {SGMRD} can handle the trade-off between computational cost and monitoring quality. 
We compare {SGMRD} to several alternative update strategies and against several baselines.
\begin{itemize}[noitemsep]
	\item \textsc{{SGMRD}-{TS}} uses MP-TS (cf. Section \ref{sec:update}) as update strategy, and we set $L=1$, unless noted otherwise. 
	\item \textsc{{SGMRD}-RD} uses a random (RD) update strategy. We update a single subspace per time step, chosen at random from the current set. 
	\item \textsc{{SGMRD}-GD} uses a greedy (GD) update strategy. We update a single subspace per time step and choose the subspace with the lowest quality. 
	\item \textsc{Batch} repeats the initialisation of {SGMRD} periodically for every batch of data with size ${w}=1000$. 
	\item \textsc{Init} runs the initialisation and then keeps the same set of subspaces for the rest of the experiment (no update). 
	\item \textsc{Gold} repeats the initialisation of {SGMRD} at every step. This baseline represents the highest level of quality that one can reach with this instantiation of {SGMRD}, but it also is the most expensive configuration. 
	In fact, we can only afford to run it on the \textsc{Pyro} data set. 
\end{itemize}

\subsubsection{Outlier Detection} 

We compare the results from \textsc{{SGMRD}-{TS}} with the following detectors: 
\begin{itemize}[noitemsep]
	\item \textsc{RS-Stream} is an adaptation of the \textsc{{RS-Hash}} \cite{DBLP:conf/icdm/SatheA16} outlier detector to the streaming setting, presented in \cite{DBLP:journals/kais/SatheA18}. It estimates the outlierness of each observation via randomised hashing over random projections. We reproduce the approach and use the default parameters recommended by the authors. 
	\item \textsc{{LOF}} \cite{DBLP:conf/sigmod/BreunigKNS00} is a well-known outlier detector. We run it periodically and average the scores over a sliding window. We use the implementation from ELKI
	\cite{DBLP:journals/pvldb/SchubertKEZSZ15}, which profits from efficient index structures.  
	\item \textsc{xStream} \cite{10.1145/3219819.3220107} is an ensemble outlier detector. \textsc{xStream} estimates densities via randomised ensemble binning  from a set of random projections. \textsc{xStream} declares the points lying in low density areas as outliers. We use the reference implementation with the recommended parameters. 
	\item \textsc{Stream{HiCS}} \cite{becker2016concept} is an adaptation of \cite{DBLP:conf/icde/KellerMB12}, repeating the initial search, based on the signals from a change detector on a data synopsis. We use the reference implementation with the recommended parameters. 
\end{itemize}

We average the scores obtained from each detector over a sliding window of size $w=1000$ for every $v=100$ time steps (cf.\ Section \ref{sec:downstream}). 
For approaches based on {LOF}, such as ours, we repeat the computation with parameter $k \in \{1,2,5,10,20,50,100\}$ and report the best result in terms of  {AUC}. 
The performance may vary widely w.r.t.\ this parameter; this is a well-known caveat of {LOF} \cite{DBLP:journals/datamine/CamposZSCMSAH16}. 
We average every result from $10$ independent runs. 
Each approach runs single-threaded on a server with 20 cores at 2.2GHz and 64GB RAM. 
We implement our algorithms in Scala.

\section{Results}
\label{sec:results}

\subsection{Subspace Monitoring}

We first evaluate the quality of monitoring from SGMRD. We set ${w}=1000$, $v=2$ and ${L}=1$, i.e., for each update strategy, {SGMRD} only keeps the latest $1000$ observations, and, for any new two observations, 
{SGMRD} attempts to replace one of the current subspaces. 

\begin{figure*}
	\includegraphics[width=\linewidth]{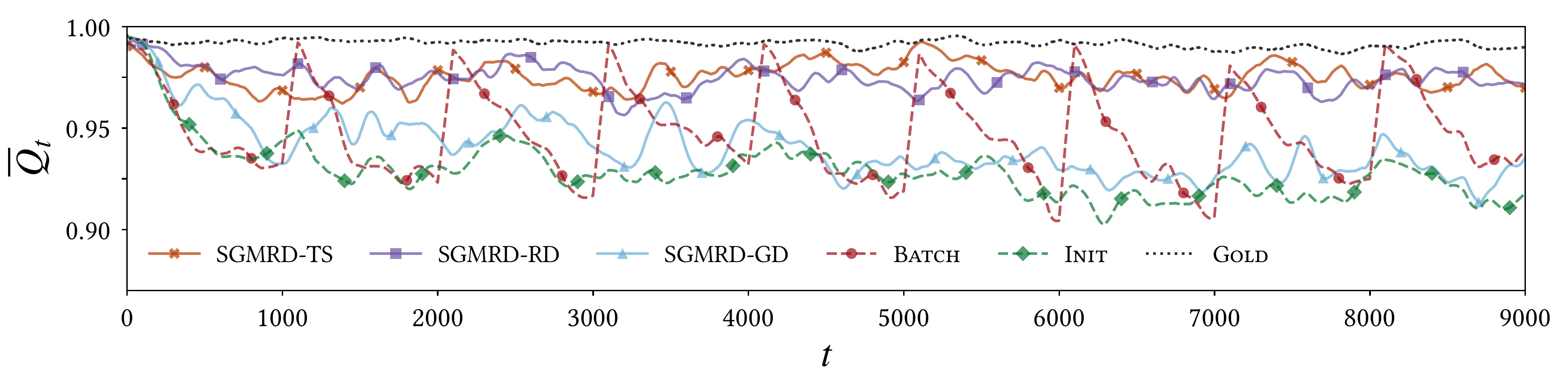}
	\caption{Average Quality at time $t$ (\textsc{Pyro}, ${L}=1$, $v = 2$).} 
	\label{fig:SGMRD_SubspaceMonitoring_Quality}
\end{figure*}

As we can see in Figure \ref{fig:SGMRD_SubspaceMonitoring_Quality}, both \textsc{{SGMRD}-{TS}} and \textsc{{SGMRD}-RD} can keep the average quality $\overline{Q}_{{t}}$ close to that of \textsc{Gold}, our strongest and most expensive baseline. 
In the beginning, \textsc{{SGMRD}-{TS}} seems to perform slightly worse than \textsc{{SGMRD}-RD}, but after some time (once \textsc{{SGMRD}-{TS}} has learned its update strategy), it tends to dominate \textsc{{SGMRD}-RD}. 
We can see that \textsc{Batch} occasionally leads to the same quality as \textsc{Gold}, but the quality drops quickly between the update steps. 
\textsc{{SGMRD}-GD} is not much better than \textsc{Init} (no monitoring). 

\begin{figure*}
	\includegraphics[width=\linewidth]{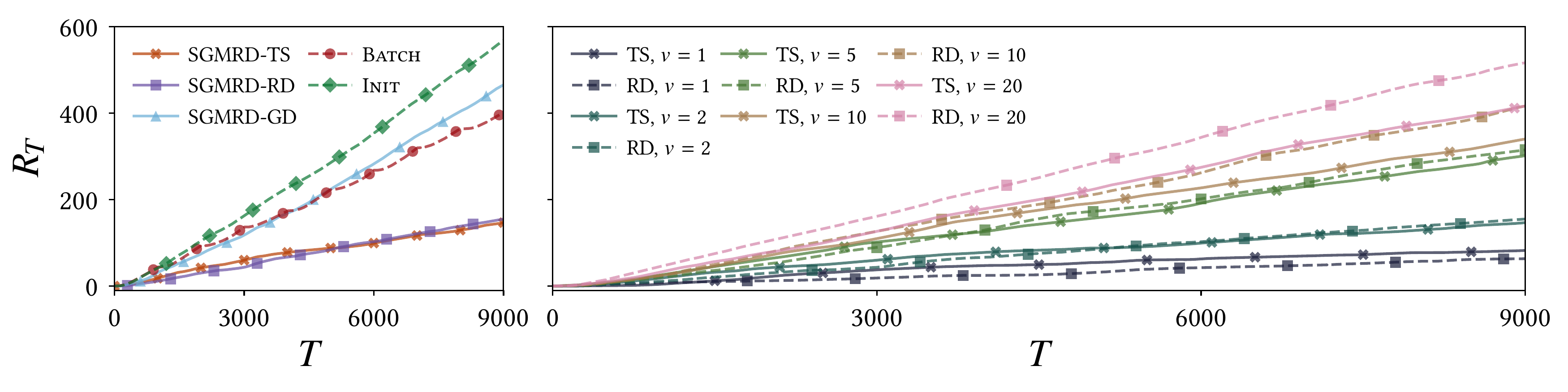}
	\caption{Regret up to $T$ (\textsc{Pyro}, ${L}=1$, left: $v = 2$).} 
	\label{fig:SGMRD_SubspaceMonitoring_Pseudoregret}
\end{figure*}

Figure \ref{fig:SGMRD_SubspaceMonitoring_Pseudoregret} confirms our observations: While the regret of \textsc{{SGMRD}-TS} is slightly worse than the one of \textsc{{SGMRD}-RD} at the beginning, it becomes better afterwards. 
The other approaches lead to much higher regret. 
For larger step size $v$, we see that \textsc{{SGMRD}-TS} is superior to \textsc{{SGMRD}-RD}. For smaller $v$, the environment does not change much between observations, and it is more difficult for bandit strategies to learn which subspaces to update more frequently. 

\begin{figure}
	\centering
	\includegraphics[width=\linewidth, trim=0 0.2cm 0 0]{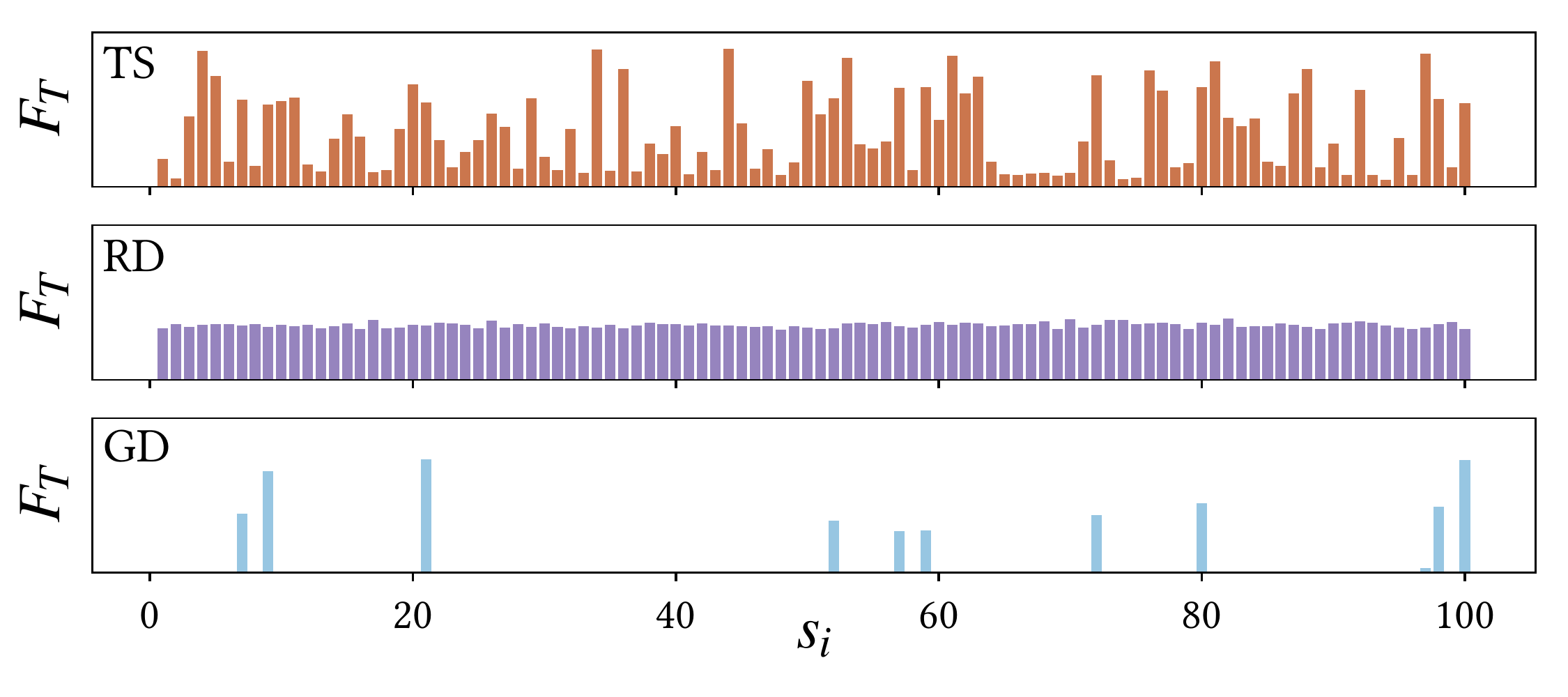}
	\caption{Frequency of update (\textsc{Pyro}, $v = 2$, $L=1$).} 
	\label{fig:SGMRD_UpdateFrequency}
\end{figure}

In Figure \ref{fig:SGMRD_UpdateFrequency}, the update frequencies give an intuition of how the three strategies differ. 
As expected, RD updates each subspace uniformly. GD tends to focus only on a few subspaces; most subspaces are never replaced, although they may become suboptimal as well. {TS} in turn tends to focus more on some subspaces, the ones requiring more frequent updates.

\begin{figure}
	\centering
	\includegraphics[width=\linewidth]{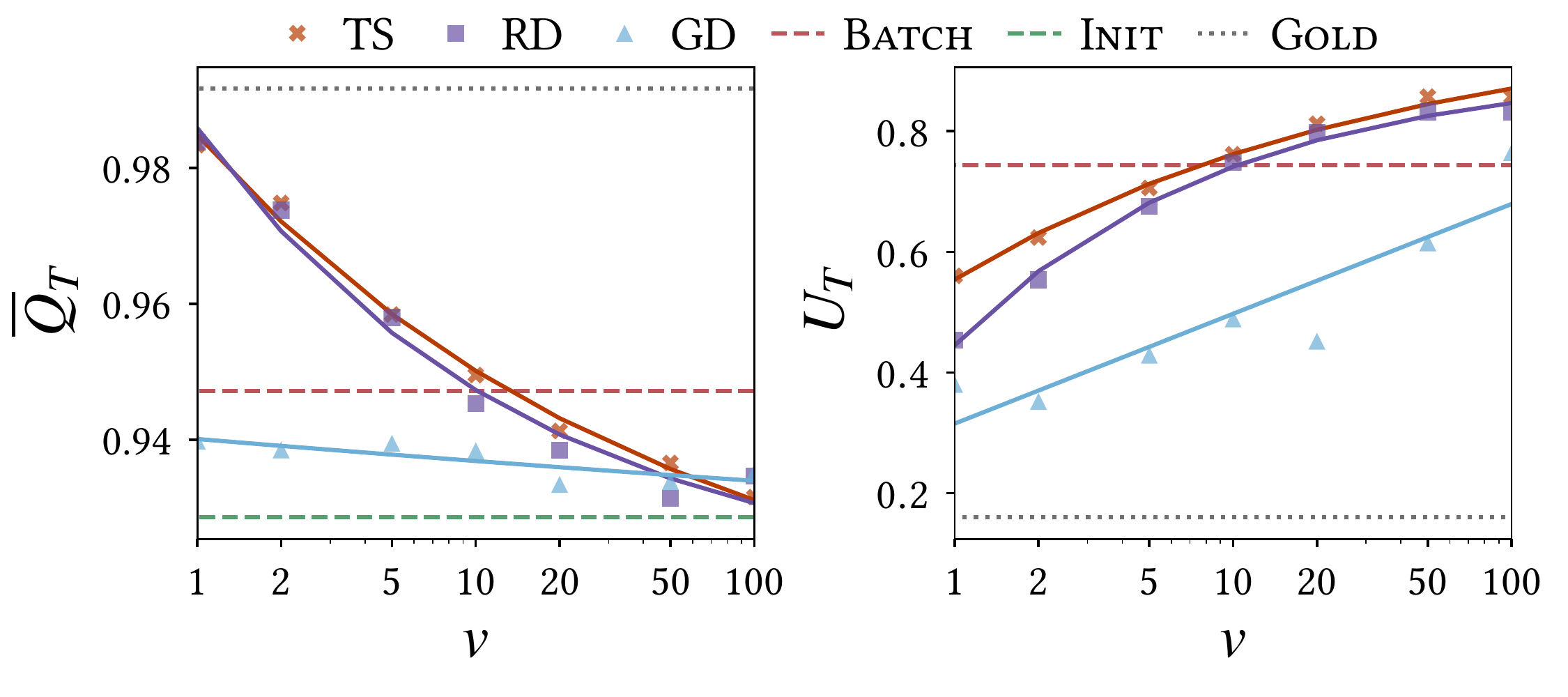}
	\caption{Quality and Success Rate w.r.t. $v$ (\textsc{Pyro}, ${L}=1$).}
	\label{fig:SGMRD_QualityUpdateWRTstep}
\end{figure} 

\begin{figure}
	\centering
	\includegraphics[width=\linewidth]{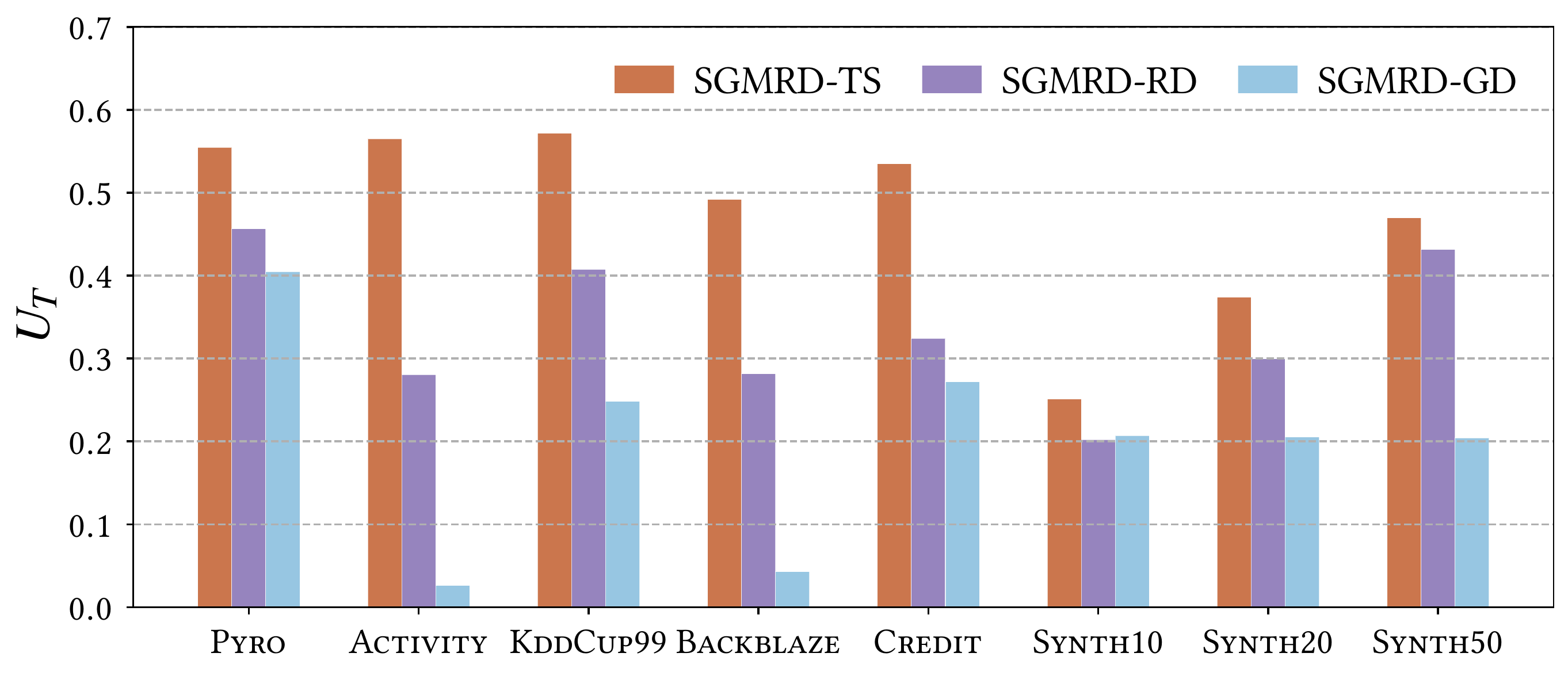}
	\caption{Success Rate (${L}=1$, $v = 1$).} 
	\label{fig:SGMRD_SuccessRatio}
\end{figure} 

\begin{figure}
	\centering
	\includegraphics[width=\linewidth]{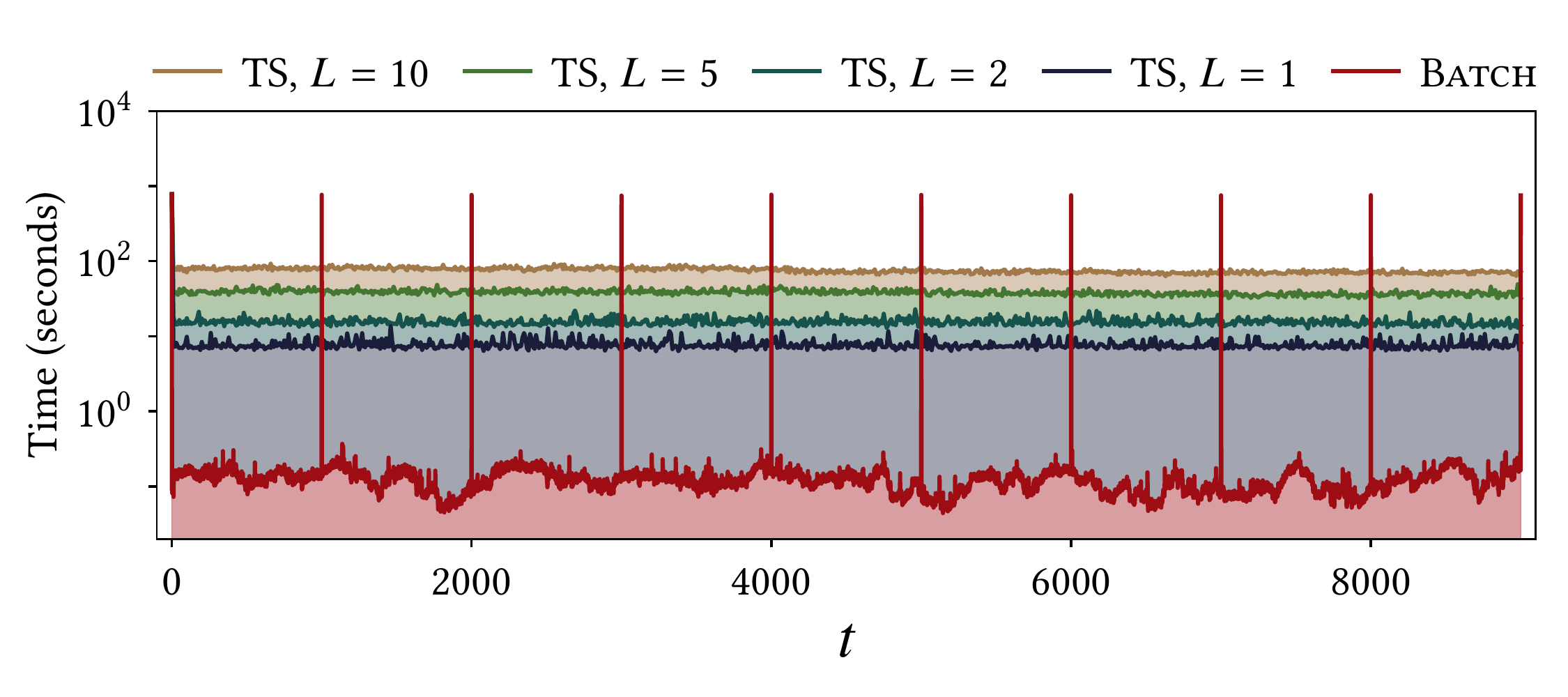}
	\caption{Stream Processing Time (\textsc{Pyro}, $v = 1$).} 
	\label{fig:SGMRD_StreamProcessing}
\end{figure}

Figure \ref{fig:SGMRD_QualityUpdateWRTstep} shows that the average quality $\overline{Q}_{{T}}$ tends to decrease as we increase the update step $v$. However, the rate of successful updates $U_{{T}}$ (Equation \ref{eq:U_T}) increases. As $v$ increases, it is more likely for any subspace to become suboptimal. 
For \textsc{Batch}, $U_{{T}}$ is high, but the quality $\overline{Q}_{{T}}$ is low. 
We observe the opposite for \textsc{Gold}. 
{SGMRD} is a trade-off between these two extremes, and the strategy based on {TS} appears superior to others, both w.r.t.\ $\overline{Q}_{{T}}$ and $U_{{T}}$. Figure \ref{fig:SGMRD_SuccessRatio} shows that our observations are not only valid for the \textsc{Pyro} data set, but also for other benchmarks. \textsc{{SGMRD}-{TS}} consistently achieves a higher rate of successful updates than other approaches.  

\begin{figure}
	\centering
	\includegraphics[width=\columnwidth]{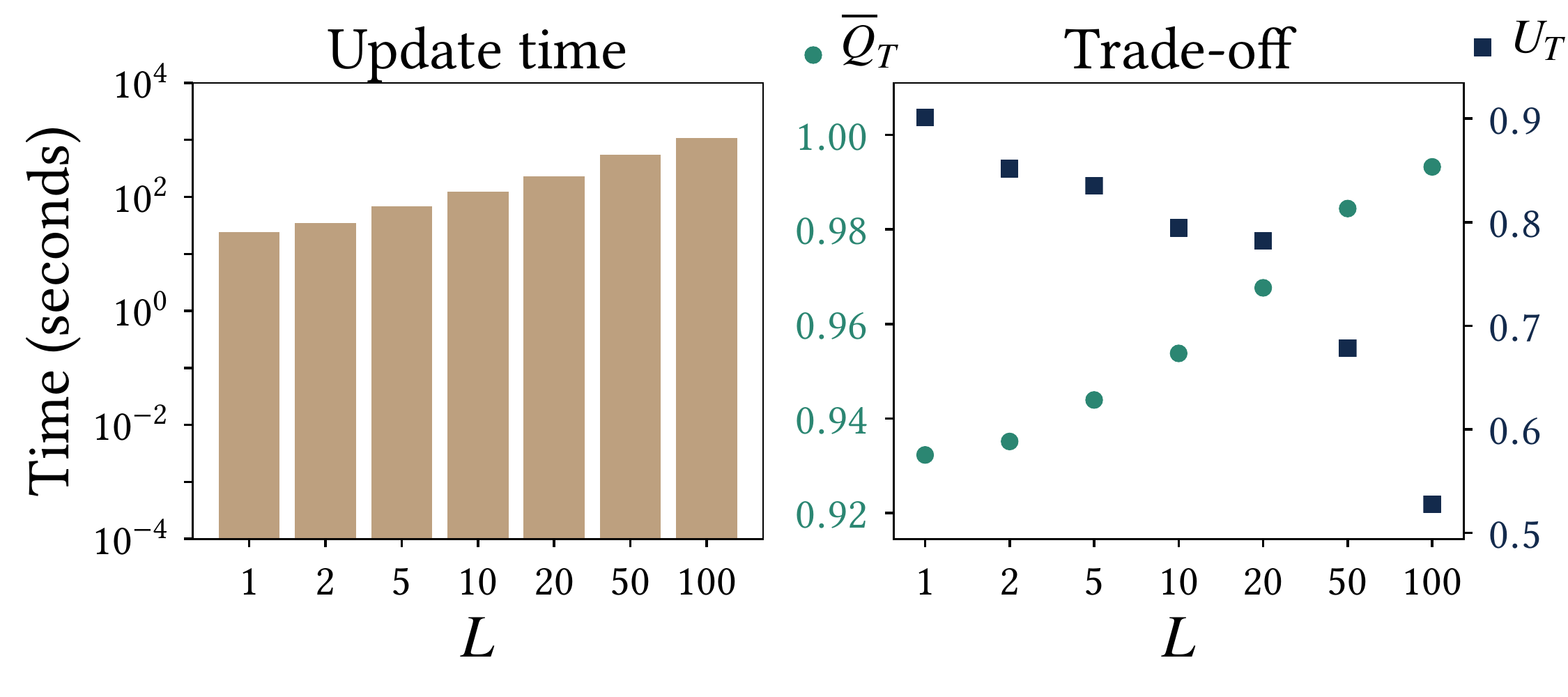}
	\caption{Quality/Efficiency (\textsc{Pyro}, \textsc{{SGMRD}-{TS}}, $v = 100$).} 
	\label{fig:SGMRD_Runtime}
\end{figure} 

Figure \ref{fig:SGMRD_StreamProcessing} highlights an important drawback of previous methods: The computation for batch-wise techniques is concentrated in a few discrete time steps. For stream mining, it is better to distribute computation uniformly over time. Computation-intensive episodes can lead to long response times of the system, and this contradicts the efficiency sought ({\textbf{C1}}) and anytime behaviour ({\textbf{C4}}). 
Besides this, the system becomes unable to adapt to the environment ({\textbf{C3}}). 

Next, we set $v=100$ and let $L$ vary to observe the trade-off between quality of the subspaces and the efficiency of the search in \textsc{{SGMRD}-{TS}} (see Figure \ref{fig:SGMRD_Runtime}). 
As ${L}$ increases, the cost of updating subspaces increases linearly. Similarly, the quality $Q_{{T}}$ increases while the rate of successful updates $U_{{T}}$ decreases.
As we can see in Table \ref{result:sgmrd_scaling}, \textsc{{SGMRD}-{TS}} has the highest quality after \textsc{Gold}, the best success rate $U_{{T}}$ after \textsc{Batch} and the smallest average regret among the baselines. 

\begin{table}
	\sisetup{detect-weight,mode=text}
	\renewrobustcmd{\bfseries}{\fontseries{b}\selectfont}
	\renewrobustcmd{\boldmath}{}
	\centering
	\caption{Comparison design alternatives (\textsc{Pyro}, $v=1$).}
	\label{result:sgmrd_scaling}
	\addtolength{\tabcolsep}{-2.5pt}
	\centering
	\small
	\renewcommand{\arraystretch}{0.82}
	\begin{tabularx}{0.8\columnwidth}{@{}lXXXX}
		\toprule
		\textbf{Baseline} & \textbf{$L_{{T}}$} & \textbf{$\overline{Q}_{{T}}$} & \textbf{$U_{{T}}$} & \textbf{$R_{{T}}/T$}  \\ \midrule 
		\textsc{SGMRD-{TS}}    & 1     & 95.43       & 0.61    &  3.74  \\
		\textsc{SGMRD-RD}    & 1           & 95.20       & 0.54    &  3.97 \\
		\textsc{SGMRD-GD}    & 1           & 93.72       & 0.43    &  5.45 \\
		\textsc{Batch} & NA          & 95.16       & 0.81    &  4.01 \\
		\textsc{Init}  & NA          & 92.50       & NA       &  6.67 \\
		\textsc{Gold}   & 100         & 99.16       & 0.16    & NA \\  \bottomrule
	\end{tabularx}
\end{table}

In conclusion, the experiments show that {SGMRD} is a useful tool to monitor high-quality subspaces over time, and it is highly versatile. 
Based on the available hardware, users can set the number of updates per round, as with \textsc{{SGMRD}-{TS}}, to obtain the highest quality for this budget. In the next section, we show that \textsc{{SGMRD}-{TS}} helps to detect outliers and compare the results with state-of-the-art outlier detectors for data streams.  

\subsection{Outlier Detection}

\begin{table}[ht]
	\sisetup{detect-weight,mode=text}
	\renewrobustcmd{\bfseries}{\fontseries{b}\selectfont}
	\renewrobustcmd{\boldmath}{}
	\centering
	\caption{Outlier Detection Performance.}
	\label{result:sgmrd_outlierdetection}
	
	\addtolength{\tabcolsep}{-4pt}
	\footnotesize
	\renewcommand{\arraystretch}{0.80}
	
	\begin{tabularx}{\columnwidth}{@{}llXXXXXXXX@{}}
		\toprule
		& Approach       & AUC            & AP             & P1\%           & P2\%           & P5\%           & R1\%           & R2\%           & R5\%           \\ \midrule
		\parbox[t]{2mm}{\multirow{5}{*}{\rotatebox[origin=c]{90}{\textsc{Activity}}}}~  & \textsc{\textbf{{SGMRD}}} & \textbf{97.32} & \textbf{85.39} & \textbf{94.59} & \textbf{94.83} & \textbf{94.24} & \textbf{9.44}  & \textbf{18.97} & \textbf{47.10} \\
		& \textsc{{LOF}}      & 93.93          & 61.80          & 74.32          & 64.72          & 64.03          & 7.42           & 12.94          & 32.00          \\
		& \textsc{Stream{HiCS}}     & 88.52          & 47.38          & 70.72          & 54.61          & 51.89          & 7.06           & 10.92          & 25.93          \\
		& \textsc{{RS-Stream}}      & 95.95          & 68.23          & 71.62          & 72.58          & 75.00          & 7.15           & 14.52          & 37.48          \\
		& \textsc{xStream}        & 77.71          & 20.41          & 3.60           & 10.14          & 16.31          & 0.36           & 2.02           & 8.13           \\ \midrule
		\parbox[t]{2mm}{\multirow{5}{*}{\rotatebox[origin=c]{90}{\textsc{KddCup99}}}}  & \textbf{{SGMRD}} & \textbf{69.98} & \textbf{10.29} & \textbf{0.00}  & \textbf{0.20}  & \textbf{0.56}  & \textbf{0.00}  & \textbf{0.06}  & \textbf{0.39}  \\
		& \textsc{LOF}      & 65.07          & 9.57           & 0.00           & 0.00           & 0.08           & 0.00           & 0.00           & 0.06           \\
		& \textsc{Stream{HiCS}}     & 57.11          & 7.89           & 0.00           & 0.00           & 0.08           & 0.00           & 0.00           & 0.06           \\
		& \textsc{{RS-Stream}}      & 43.21          & 5.73           & 0.00           & 0.00           & 0.08           & 0.00           & 0.00           & 0.06           \\
		& \textsc{xStream}        & 52.70          & 8.23           & 0.00           & 0.20           & 0.08           & 0.00           & 0.06           & 0.06           \\ \midrule
		\parbox[t]{2mm}{\multirow{5}{*}{\rotatebox[origin=c]{90}{\textsc{Backblaze}}}}  & \textbf{{SGMRD}} & \textbf{90.91} &	13.31  & 	7.14 & 	18.65 &	\textbf{15.40} & 	7.14 &	37.30 &	\textbf{76.98}  \\
		& \textsc{LOF}      & 56.92 &	1.65 &	2.38 &	3.57 &	2.38 &	2.38 &	7.14 &	11.90           \\
		& \textsc{Stream{HiCS}}     & 79.22 &	\textbf{40.07} &	\textbf{50.79} &	\textbf{26.59} &	10.95 &	\textbf{50.79} &	\textbf{53.17} &	54.76          \\
		& \textsc{{RS-Stream}}      & 80.55 &	7.17 &	7.14 &	13.49 &	9.37 &	7.14 &	26.98 &	46.83         \\
		& \textsc{xStream}        & 76.86        & 3.69           &  1.59 & 3.17 &  6.19  &  1.59 & 6.35 & 30.95         \\ \midrule
		\parbox[t]{2mm}{\multirow{5}{*}{\rotatebox[origin=c]{90}{\textsc{Credit}}}}  & \textbf{{SGMRD}} & \textbf{95.06} & 	\textbf{15.87} & 	10.69 &	\textbf{7.47} &	\textbf{3.22} &	55.51 &	\textbf{77.57} &	\textbf{83.65}  \\
		& \textsc{LOF}      & 91.50 &	4.67 &	6.22 &	5.20 &	2.69 &	32.32 &	53.99 &	69.96          \\
		& \textsc{Stream{HiCS}}     & 89.21 &	3.48 &	3.59 &	2.93 &	2.28 &	18.63 &	30.42 &	59.32         \\
		& \textsc{{RS-Stream}}      & 85.13 &	1.63 &	2.27 &	2.49 &	1.86 &	11.79 &	25.86 &	48.29          \\
		& \textsc{xStream}        &  94.62 &  9.10  & \textbf{10.83} & 6.48 & 3.18 & \textbf{56.27} & 67.30 & 82.51   \\ \midrule
		\parbox[t]{2mm}{\multirow{5}{*}{\rotatebox[origin=c]{90}{\textsc{Synth10}}}}   & \textbf{{SGMRD}} & \textbf{92.70} & \textbf{59.93} & \textbf{50.00} & \textbf{26.00} & \textbf{12.00} & \textbf{58.14} & \textbf{60.47} & \textbf{69.77} \\
		& \textsc{{LOF}}      & 88.77          & 31.44          & 33.00          & 18.50          & 10.40          & 38.37          & 43.02          & 60.47          \\
		& \textsc{Stream{HiCS}}     & 88.81          & 31.16          & 33.00          & 19.00          & 10.40          & 38.37          & 44.19          & 60.47          \\
		& \textsc{{RS-Stream}}      & 71.23          & 1.87           & 0.00           & 0.00           & 2.80           & 0.00           & 0.00           & 16.28          \\
		& \textsc{xStream}        & 68.51          & 2.58           & 5.00           & 3.00           & 4.00           & 5.81           & 6.98           & 23.26          \\ \midrule
		\parbox[t]{2mm}{\multirow{5}{*}{\rotatebox[origin=c]{90}{\textsc{Synth20}}}}  & \textbf{{SGMRD}} & \textbf{85.05} & \textbf{41.19} & \textbf{36.00} & \textbf{19.50} & \textbf{9.20}  & \textbf{40.91} & \textbf{44.32} & \textbf{52.27} \\
		& \textsc{{LOF}}      & 72.55          & 5.57           & 8.00           & 6.00           & 4.40           & 9.09           & 13.64          & 25.00          \\
		& \textsc{Stream{HiCS}}     & 71.71          & 5.37           & 8.00           & 6.00           & 4.00           & 9.09           & 13.64          & 22.73          \\
		& \textsc{{RS-Stream}}      & 48.39          & 0.80           & 0.00           & 0.00           & 0.00           & 0.00           & 0.00           & 0.00           \\
		& \textsc{xStream}        & 63.64          & 1.58           & 1.00           & 1.50           & 2.20           & 1.14           & 3.41           & 12.50          \\ \midrule
		\parbox[t]{2mm}{\multirow{5}{*}{\rotatebox[origin=c]{90}{\textsc{Synth50}}}}   & \textbf{{SGMRD}} & \textbf{75.87} & \textbf{31.27} & \textbf{27.00} & \textbf{16.00} & \textbf{7.60}  & \textbf{33.33} & \textbf{39.51} & \textbf{46.91} \\
		& \textsc{{LOF}}      & 61.38          & 1.08           & 0.00           & 0.50           & 0.60           & 0.00           & 1.23           & 3.70           \\
		& \textsc{Stream{HiCS}}     & 63.90          & 12.00          & 11.00          & 6.00           & 3.40           & 13.58          & 14.81          & 20.99          \\
		& \textsc{{RS-Stream}}      & 46.52          & 0.73           & 0.00           & 0.00           & 0.00           & 0.00           & 0.00           & 0.00           \\
		& \textsc{xStream}        & 48.43          & 0.90           & 1.00           & 0.50           & 1.40           & 1.23           & 1.23           & 8.64           \\ \bottomrule
	\end{tabularx}
	\vspace{-0.5cm}
\end{table}

\begin{figure}
	\centering
	\includegraphics[width=\linewidth]{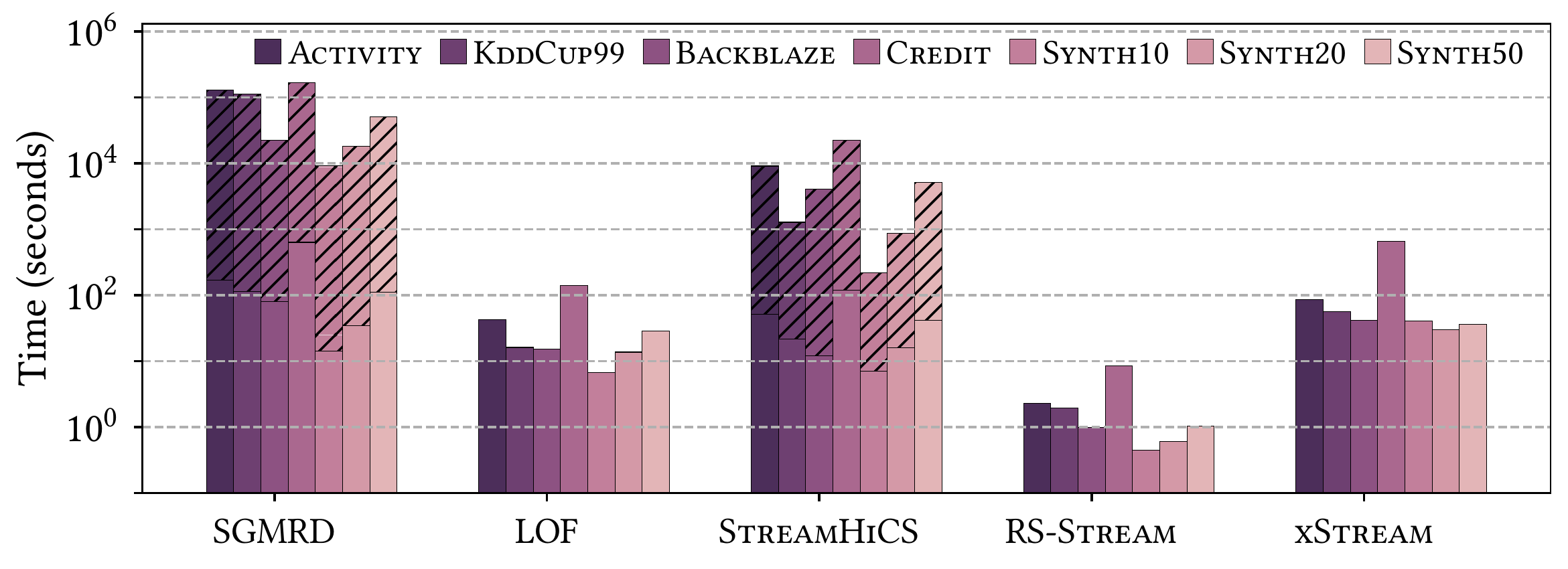}
	\caption{Outlier Detection Time (hatched: search time).} 
	\label{fig:SGMRD_OutlierDetectionRuntime}
	\vspace{-0.2cm}
\end{figure}

We leverage the subspaces obtained from \textsc{{SGMRD}-{TS}} to detect outliers, as in Section \ref{sec:downstream}. We set ${w}=1000$, $v=1$ and ${L}=1$. Table \ref{result:sgmrd_outlierdetection} shows the results. Compared to our competitors, \textsc{{SGMRD}} clearly leads to the best results w.r.t.\ each benchmark. Interestingly, \textsc{StreamHiCS} appears to have higher recall and precision in the top 1-2\%. We can see that {LOF} turns out to be our most competitive baseline and often outperforms our competitors. 

In Figure \ref{fig:SGMRD_OutlierDetectionRuntime}, we can see that our competitors, in particular \textsc{{RS-Stream}} and \textsc{xStream}, are much faster than \textsc{{SGMRD}}, but they often are not much better than random guessing. 
Most of the computation required by \textsc{{SGMRD}} and \textsc{Stream{HiCS}} is due to the search. 
Nonetheless, one may reduce the required computation with \textsc{{SGMRD}}, e.g., increase $v$, without decreasing detection quality by much. 

\section{Conclusions}
\label{sec:conclusions}

Finding interesting subspaces is fundamental to any step of the knowledge discovery process. We have proposed a new method, {SGMRD}, to bring subspace search to streams. It does so by combining an efficient greedy heuristic with novel multivariate quality estimators and an efficient bandit-based strategy, to update the results of subspace search over time.  

Our experiments not only show that {SGMRD} leads to efficient monitoring of subspaces, but also to state-of-the-art results w.r.t.\ downstream data mining tasks, such as outlier detection, and one may expect similar benefits for other mining tasks on data streams. 

An administrator can control monitoring via two parameters: the number of plays per round $L$ and the step size $v$. 
While the impact of a parameter value can be studied empirically, finding the most adequate parameters for a specific problem is not trivial. 
In future work, it would be interesting to extend our update policy in order to automatically tune those parameters, and to deal with their potential non-stationarity in the streaming setting. A possible direction is to leverage the recent bandit models from \cite{DBLP:conf/kdd/FoucheKB19}. 

\begin{acks}
This work was supported by the DFG Research Training Group 2153: ``Energy Status Data -- Informatics Methods for its Collection, Analysis and Exploitation'' and the German Federal Ministry of Education and Research (BMBF) via Software Campus (01IS17042). 
\end{acks}

\bibliographystyle{ACM-Reference-Format}
\bibliography{sgmrd}

\typeout{get arXiv to do 4 passes: Label(s) may have changed. Rerun} 
\end{document}